\documentclass[letterpaper, 10 pt, conference]{IEEEtran} 
\IEEEoverridecommandlockouts                              % This command is only needed if 
\title{\LARGE \bf
Should ChatGPT and Bard Share Revenue with Their Data Providers? A New Business Model for the AI Era 
}

\author{Dong Zhang $^{1}$\thanks{$^{1}$ {\tt\small dongzhanghz@gmail.com}. The content of the paper is based on the author's own work, which does not reflect the views and opinions of the author's employer.}
        }
\usepackage{float}
\usepackage{graphicx}
\usepackage{comment}
\usepackage{amsmath,amssymb} % define this before the line numbering.
\usepackage{color}
\usepackage{algpseudocode}
\usepackage{caption}
\usepackage{subcaption}
\usepackage{mathtools}
\usepackage{caption}
\usepackage[
  separate-uncertainty = true,
  multi-part-units = repeat
]{siunitx}
\usepackage{subcaption}
\usepackage{float}

\definecolor{red}{rgb}{1.00,0.00,0.00}
\definecolor{blue}{rgb}{0.00,0.00,1.00}
\definecolor{green}{rgb}{0.2,0.70,0.2}
\definecolor{yellow}{rgb}{0.5,0.5,0.0}

\usepackage{graphicx}
\usepackage{amsmath,amssymb,amsfonts}
\usepackage{hyperref}
\usepackage[ruled,vlined]{algorithm2e}
\usepackage{multirow}
\usepackage{tabularx, booktabs}
\usepackage{mathtools}
\usepackage{arydshln}
\usepackage{xcolor}
\usepackage{mwe}
\usepackage{wrapfig}
\usepackage[bottom]{footmisc} % put footnote under figures
\usepackage{hyperref}
\usepackage{makecell}
\usepackage{todonotes}
\usepackage{mathrsfs}

\usepackage{listings}

\begin{document}

\maketitle
\thispagestyle{plain}
\pagestyle{plain}

%%%%%%%%%%%%%%%%%%%%%%%%%%%%%%%%%%%%%%%%%%%%%%%%%%%%%%%%%%%%%%%%%%%%%%%%%%%%%%%%
\begin{abstract}

With various AI tools such as ChatGPT becoming increasingly convenient and popular, we are entering a true AI era. We can foresee that exceptional AI tools will soon reap considerable profits. A crucial question arise: should AI tools share revenue with their training data providers in additional to traditional stakeholders and shareholders? The answer is Yes. Large AI tools, such as large language models, always require more and better quality data to continuously improve, but current copyright laws limit their access to various types of data. Sharing revenue between AI tools and their data providers could transform the current hostile zero-sum game relationship between AI tools and a majority of copyrighted data owners into a collaborative and mutually beneficial one, which is necessary to facilitate the development of a virtuous cycle among AI tools, their users and data providers that drives forward AI technology and builds a healthy AI ecosystem. However, current revenue-sharing business models do not work for AI tools in the forthcoming AI era, since the most widely used metrics for website-based traffic and action, such as clicks, will be replaced by new metrics such as prompts and cost per prompt for generative AI tools. Therefore, a completely new revenue-sharing business model must be established. This new business model, which must be almost independent of AI tools and be easily explained to data providers, needs to establish a prompt-based scoring system to measure data engagement of each data provider. This paper systematically discusses how to build such a scoring system for all data providers for AI tools based on classification and content similarity models, and outlines the requirements for AI tools or third parties to build it. AI tools can share revenue with data providers using such a scoring system, which would encourage more data owners to participate in the revenue-sharing program. This will be a utilitarian AI era where all parties benefit.

\end{abstract}

%%%%%%%%%%%%%%%%%%%%%%%%%%%%%%%%%%%%%%%%%%%%%%%%%%%%%%%%%%%%%%%%%%%%%%%%%%%%%%%%

\setcounter{tocdepth}{2}
\tableofcontents

\section{Introduction}\label{section_intro}

ChatGPT (Chat Generative Pre-trained Transformer) has recently become highly popular, and its popularity is continuing to rise. It is a class of advanced machine learning Natural Language Processing models developed by OpenAI, using both supervised machine learning and reinforcement learning techniques. ChatGPT was initially built on top of OpenAI's GPT-3 family, then later GPT-4, designed to answer questions, generate text, summarize information, and perform other language-related tasks. It has the capability to understand and respond to various types of inputs, making it a powerful tool for conversational Artificial Intelligence (AI) applications.

As a large language model (LLM), ChatGPT is trained on a large and diverse dataset of text obtained from various sources on the internet, but the specific sources and details of the training data are still not publicly available. Despite this, we can still gain some understanding from its previous version GPT-3 \cite{GPT3_2020}. As of 2020, GPT-3 used a broad range of data sources, including the CommonCrawl data, WebText, two internet-based books corpora, as well as English-language Wikipedia. The \href{https://commoncrawl.org/}{CommonCrawl} data contains 410 billion tokens originally from billions of web pages collected by the CommonCrawl Foundation. The WebText2 dataset, which comes from high quality webpages scraped from Reddit links, contains over 19 billon tokens. The two book sources include about 67 billion tokens, while Wikipedia contributes at least 3 billion tokens. GPT-3 API was opened to the public in 2021, and InstructGPT, which used reinforcement learning from human feedback (RLHF) to fine-tune GPT-3, was released in 2021 and upgraded in 2022 \cite{InstructGPT}. ChatGPT, which is a fine-tuned version of a model from the GPT-3.5 series and a sibling model to InstructGPT, trained to respond to a prompt with a detailed response.  

ChatGPT was made available for public testing on November 30th, 2022. Just one week after the launch, Samuel Altman, CEO of OpenAI announced on \href{https://twitter.com/sama/status/1599668808285028353}{Twitter} that ChatGPT has crossed one million users. It was reported that ChatGPT has more than 100 million users with the first two months of its launch, and has more than 13 million daily visitors as of 2023. On February 1, OpenAI announced that it is rolling out \href{https://openai.com/blog/chatgpt-plus/}{ChatGPT Plus} pilot plan, a premium subscription service for \$20 per month, first in the US marketplace, then worldwide, to have faster ChatGPT access and response times, as well as priority access to new ChatGPT features and improvements. On March 14, OpenAI launched \href{https://openai.com/research/gpt-4}{GPT-4}, which limited version became public available in ChatGPT Plus. OpenAI predicts that ChatGPT will generate \$200 million in revenue by the end of 2023, and \$ 1 billion in 2024. Of course OpenAI will distribute its revenue from ChatGPT to its stakeholders. For example, Microsoft will reap significant profits from OpenAI, since Microsoft Azure has supported OpenAI and ChatGPT model training, and it has invested \$10 billion in OpenAI\footnote{See report by \href{https://www.reuters.com/business/chatgpt-owner-openai-projects-1-billion-revenue-by-2024-sources-2022-12-15/}{Reuters}.}. It was reported that Microsoft may gain 75\% share of OpenAI’s profits until it makes back the money on its investment, after which 49\% of stake ownership of it\footnote{See news from \href{https://www.cnbc.com/2023/01/10/microsoft-to-invest-10-billion-in-chatgpt-creator-openai-report-says.html}{CNBC} and \href{https://www.reuters.com/technology/microsoft-talks-invest-10-bln-chatgpt-owner-semafor-2023-01-10/}{Reuters}.}. Meanwhile, Google also launched its own chatbot \href{https://bard.google.com}{Bard} publicly in March 2023, other publicly available chatbots based on LLMs launched almost at the same time including Anthropic's \href{https://www.anthropic.com/index/introducing-claude}{Claude}, Stanford's \href{https://crfm.stanford.edu/2023/03/13/alpaca.html}{Alpaca}, Baidu's \href{https://yiyan.baidu.com}{ERNIE Bot}, Databricks' \href{https://www.databricks.com/blog/2023/03/24/hello-dolly-democratizing-magic-chatgpt-open-models.html}{Dolly}, Alibaba's \href{https://tongyi.aliyun.comand}{Tongyi Qianwen}, and so on\footnote{In the title of this paper, I only mentioned ChatGPT and Bard as representatives of LLM-based chatbots and even AI tools. For a more detailed list of LLMs, please refer to Section \ref{section_utili}.}. 

%Microsoft's \href{https://github.com/microsoft/JARVIS}{JARVIS}, Databricks' \href{https://www.databricks.com/blog/2023/03/24/hello-dolly-democratizing-magic-chatgpt-open-models.html}{Dolly}

But there is a question here. Using ChatGPT as an example, should OpenAI only share the revenue and profits from ChatGPT with its stakeholders and shareholders, or should it also share them with all the data providers involved? We know that OpenAI has expended a great deal of effort collecting the training dataset for its GPT families, and has consumed a significant amount of time developing and training the models with semi-supervised, supervised and RLHF learnings, but the success of ChatGPT is largely due not only to its state-of-the-art models, but also to the huge amount of data it has collected and used for training. Would it not be fair to acknowledge and credit the contribution of all the providers of ChatGPT's diverse data sources to its success? It is worth noting that even if the data sources are in public domain, it is still common practice to give proper attribution when using them in one's work, so should be ChatGPT. But how can an AI tool weight the importance of its data providers and acknowledge them? In this paper, we will see that discussing these questions is necessary and essential for building a good AI ecosystem.

Of course the above questions are not only for ChatGPT, or for LLMs, but for almost all AI tools that are emerging. We have AI tools in various fields. For example, OpenAI also has a popular image generator called \href{https://openai.com/product/dall-e-2}{Dall-E 2}. Similar AI image generators including \href{https://stability.ai/}{Stability AI}'s \href{https://github.com/CompVis/stable-diffusion}{Stable Diffusion}, \href{https://www.midjourney.com}{MidJourney}, \href{https://starryai.com}{Starryai} and so on. Although the training datasets of most AI image generators are still not yet public, we know that at least they have been using \href{https://laion.ai/projects}{LAION} (Large-scale Artificial Intelligence Open Network) image datasets, which has more than 5 billion images, paired with text captions.  Should the commercialized AI image generators share revenue with LAION, even LAION is a non-profit organization? Or should they also share revenue with artists and photographers who originally created or made the images? Most AI image generators' companies have detailed their compliance with Digital Millennium Copyright Act (DMCA) on their official websites, with the aim of avoiding any infringement of someone's intellectual property rights. But some artists have initiated lawsuits against these three image generator companies since they believe that these companies have violated the DMCA\footnote{Lawsuit against some AI image generators was reported by multiple media outlets, such as \href{https://www.cbsnews.com/news/ai-stable-diffusion-stability-ai-lawsuit-artists-sue-image-generators/}{CBS News}, \href{https://www.theverge.com/2023/1/16/23557098/generative-ai-art-copyright-legal-lawsuit-stable-diffusion-midjourney-deviantart}{The Verge}, \href{https://news.bloomberglaw.com/ip-law/first-ai-art-generator-lawsuits-threaten-future-of-emerging-tech}{Bloomberg}. More lawsuits see Section \ref{section_text2image}.}. In the forthcoming AI era, is it better for AI tools to strictly adhere to existing copyright laws such as DMCA, or is sharing revenue with data owners, such as artists, photographers, writers, a mutually beneficial better approach? Let us take another look at the field of healthcare. Medical imaging is another area that utilizes AI tools. Public and private medical imaging datasets are used to compare with images uploaded by customers on health platforms such as Google Cloud's  \href{https://cloud.google.com/medical-imaging}{Medical Imaging Suite} or Amazon Web Service (AWS)'s \href{https://aws.amazon.com/healthlake/}{HealthLake}, so that customers can do modeling and data analyze for their own images. But so far AI tools for healthcare still lack sufficient data. Can we create a new mechanism where health platforms share revenue with the providers of the medical imaging data?  And this new mechanism may also encourage more medical institutions or personals to join and share their medical data, and eventually significantly improve the health AI tools. 

%Although the training dataset of Dall-E 2 is still not yet public, we know that Dall-E 2 at least used Google's Conceptual Captions dataset, which has more than 3 million images, paired with natural-language captions. Should the commercialized Dall-E 2 share the revenue with Google, or even the original image creators from Conceptual Captions? 

Currently, there are multiple ways for companies to distribute revenue with their content providers. Some companies pay contributors through a royalty-based system, where contributors earn based on the usage of their content and the licensing agreement between them and companies. For example, Getty images contributors can earn 20\% and 25\% of consumers' payment for each image and video respectively, while iStock exclusive contributors can earn between 15\% and 45\% loyalties for each piece of content they license\footnote{The rates are from \href{https://photutorial.com/istock-and-getty-images-contributor-review}{iStock and Getty Images Contributor Earnings}.}.  However, this method of revenue sharing is more conventional and may not be suitable for the emerging AI industry.  In the Web 2.0 era, a more sophisticated and internet way to share revenue is based on cost per action (CPA). Google has developed a revenue sharing mechanism through a platform called AdSense, which is an advertising platform that enables website owners to display targeted ads on their websites and earn money for every click on those ads. This sharing is based on some CPA models, which are popular today, but not good enough for the new coming AI era. It is very likely that we can have a more ``AI way" to ask emerging  big AI tech companies to share their revenue, and to encourage more people and organizations to participate in the AI era. 

This paper provides a systematic discussion on how to establish a revenue-sharing business model for AI tools. It is organized as follows. Section \ref{shareRev_good} discusses why sharing revenue is not only a good idea for AI tools which want to grow, but also for the entire AI ecosystem in the upcoming AI era. Section \ref{section_today_model} reviews current revenue sharing models, such as Getty image's and Google's model, and argues that new metrics and new business model are needed for AI tools and their data providers. A scoring system for data providers is proposed in Section \ref{revenue_share_model_AI}, and multiple methods including supervised classification and content similarity calculations for building such a system are demonstrated and discussed, which can be applied to any AI tool in general. Section \ref{section_discussion} explores how such a business model can be applied to AI image generators, AI in healthcare, and multimodal tools. Finally, Section \ref{section_conclusion} presents the conclusion and proposed future work.

\section{Why Sharing Revenue is a Good Idea}\label{shareRev_good}

In the following discussion, I will often refer to ChatGPT. But I am not just targeting ChatGPT, which is just an example. The discussion in this paper is essentially applicable to most large commercialized  AI tools.

\subsection{Data vs. Models}\label{section_data_model}

Deep learning models used in AI tools rely heavily on large amounts of data to improve their accuracy and performance. Without data, AI tools cannot make any progress. Peter Norvig, Google's research director once said, ``We don’t have better algorithms. We just have more data. More data beats clever algorithm, but better data beats more data." Although there is still debate about whether data or machine learning algorithms are more important for the coming AI era, it is undeniable that machine learning especially deep learning model performance highly depends on the quantity and quality of its training data \cite{more_data_2009, more_data_2017}. This can also be seen from the evolution of GPT models. The first generation model, GPT-1 \cite{GPT1_2018}, was trained on the BooksCorpus dataset of 4.5\,GB with about 120 million parameters \cite{Bookscropus15}. The second generation model, GPT-2 \cite{GPT2_2019}, used a 40\,GB of \href{https://paperswithcode.com/dataset/webtext}{WebText} training dataset to feed a larger architecture with the number of parameters increasing to 1.5 billion. The following GPT-3 had a 45\,TB training dataset before preprocessing across five different corpora including Common Crawl, WebText2, Books1, Books2 and Wikipedia, feeding an even larger architecture with 175 billion parameters \cite{GPT3_2020}. The training dataset for GPT-4 is still a giant black box, but it is very likely significantly larger than GPT-3's training dataset. If we consider that ChatGPT and other LLMs sparked an AI revolution, then this revolution is undoubtedly supported by massive amount of data.

\subsection{Still need more and better data}\label{section_more_data}

%However, more data does not always lead to better model performance.

But ChatGPT is not perfect. Contrary to what many people imagine, the current generation of ChatGPT has many limitations. One of the major problems is that many of the answers it gives are incorrect. ChatGPT itself acknowledged that it ``sometimes writes plausible-sounding but incorrect or nonsensical answers". The question and answer website Stack Overflow banned the use of ChatGPT for generating answers to questions, since ChatGPT ``produces have a high rate of being incorrect"\footnote{See this Stack Overflow \href{https://meta.stackoverflow.com/questions/421831/temporary-policy-chatgpt-is-banned}{announcement} for more details.}. ChatGPT also generates fake scientific abstracts and research papers\footnote{See Forbes news \href{https://www.forbes.com/sites/brianbushard/2023/01/10/fake-scientific-abstracts-written-by-chatgpt-fooled-scientists-study-finds}{Fake Scientific Abstracts Written By ChatGPT Fooled Scientists, Study Finds} (10 Jan 2023) and Nature news \href{https://www.nature.com/articles/d41586-023-00056-7}{Abstracts written by ChatGPT fool scientists} (12 Jan 2023)}, making the academic community very cautious about using it. Not only ChatGPT, but also other advanced chatbots made errors. This is called AI ``hallucination problem". Meta's Galactica model \cite{Galactica}, which was trained on 48 million examples of a variety of sources, was offline after experts found it to be biased and generating false information. Microsoft's GPT-powered new Bing made mistakes in demo, and Bard, Google's ChatGPT competitor, gave wrong answer related to James Webb Space Telescope in an initial promotional demo.   

One way to address the limitations of today's chatbots is to restrict their use cases, such as making ChatGPT only an auxiliary tool for writing, drafting documents, summarizing meeting records and giving limited recommendation. But this is clearly not what we want to see in the coming AI era, because even today's digital assistant tools such as Siri and Alexa can provide more information services. The ChatGPT website discusses the reasons for its incorrect answers, in addition to model-related reasons, a major factor is the lack of ground truth data. Obviously, the next generation of ChatGPT needs more and better data to train a better version. Here, \textbf{``more and better data"} not only means having a large training dataset for the model (just as some paper discussed \cite{more_data_2009, more_data_2017}), but also has at least four manifold meanings: 

\begin{itemize}
\item[1.] Enhancing the diversity and amount of training data and reducing data noise can improve models' learning ability and decrease the chances of errors. 
\item[2.] Using data specific to a certain task of domain knowledge can fine tune the model for that task or domain. 
\item[3.] Many models, such as ChatGPT, are using outdated data. It is important to keep the data up-to-date to produce more precise and recent results. 
\item[4.] Incorporating more human feedback to establish better ground truth. So far chatGPT has done better job to collect human feedback than other LLMs. As competitive products like Bard gather a large amount of user feedback, we will see which chatbots and LLMs can better utilize this feedback.

\end{itemize}

So far, many large AI tools such as ChatGPT only use training data from publicly available sources, even so they have faced controversy. Italy has recently imposed a temporary ban on ChatGPT due to concerns over potential violations of European privacy regulations. Definitely not all important data is publicly available, and ChatGPT, as well as any AI tool that wants to be sufficient intelligent, must have the willingness and methods to purchase data from diverse sources. But how can an AI tool purchase data outside of the public domain without violating regulations? Before discussing this topic, let us take a look at data ownership and copyrights. 

%all the data used by large AI tools such as ChatGPT comes from publicly available sources (there have been some lawsuits but the situation is tricky). However, not all important data is publicly available, and ChatGPT, as well as any AI tool that wants to be sufficient intelligent, must have the willingness and methods to purchase data from diverse sources. But how can an AI tool purchase data outside of the public domain? Before discussing this topic, let us take a look at data ownership and copyrights. 

\subsection{Data Privacy and Copyright}\label{section_copyright}

Can ChatGPT use recent articles from New York Times, Wall Street Journal or BBC for training? I am afraid not. Most of the articles from these media outlets are not legally available for use without permission. The robots and crawlers policies of many media do not allow for free scraping of their content. Although whether media outlets used for training are not publicly disclosed by OpenAI, but there are now some concerns that ChatGPT may have used copyrighted materials from certain media outlets\footnote{For example, Bloomberg's article \href{https://www.bloomberg.com/news/articles/2023-02-17/openai-is-faulted-by-media-for-using-articles-to-train-chatgpt}{OpenAI Is Faulted by Media for Using Articles to Train ChatGPT} (Feb 16 2023).}. If that is the case, ChatGPT may face legal challenges,  this is certainly something that ChatGPT wants to avoid as much as possible. Since it is too expensive to use traditional way to purchase copyrighted data, this has resulted in many important news knowledge being excluded from the current training corpus of ChatGPT. Other AI tools also face similar issues.

Moreover, books are invaluable for long-range context modeling and coherent storytelling. ChatGTP has also used a huge number of books for training, but book selection is limited to public domain only. We know that the \href{https://www.gutenberg.org}{Project Gutenberg} is a library of over 60,000 free books with more than 10\,GB data and 3 billion tokens, GPT-3 used Books1 and Books2 datasets with a total of over 60 billion tokens. However, keep in mind there are $\sim$\,130 million books published since the invention of Gutenberg's printing press in 1440 A.D.\footnote{This number was estimated by \href{http://booksearch.blogspot.com/2010/08/books-of-world-stand-up-and-be-counted.html}{Google} a decade ago.}, a majority of books are copyrighted. Should ChatGPT use a traditional way to spend \$10 or \$20 to buy each book to feed the model? Absolutely not.  

ChatGTP also uses Wikipedia and gives it a high weight in the training dataset to emphasize its importance. However, Wikipedia is not a reliable source for citations, as it states itself: ``As a user-generated source, it can be edited by anyone at any time, and any information it contains at a particular time could be vandalism, a work in progress, or simply incorrect."\footnote{See Wikipeadia's own statement: \href{https://en.wikipedia.org/wiki/Wikipedia:Wikipedia_is_not_a_reliable_source}{Wikipedia is not a reliable source}.} In contrast to Wikipedia, Encyclopedia Britannica is a well-known, centuries-old English-language encyclopedia that seems to have a reputation for scholarly authority. Why cannot ChatGPT use Encyclopedia Britannica or other encyclopedias as a more reliable source? Again, the reason is that other encyclopedias are copyrighted sources and may not be available for unrestricted use. Even if ChatGPT buys some encyclopedias, it still cannot freely display the knowledge from them to the public, according to current copyright laws. 

To summarize, in the current situation, ChatGPT can only use public domain data for training, but it is obvious that public domain data is not enough. Just like a student needs to buy some textbooks, if ChatGPT wants to become an excellent student, it must spend money to purchase "textbooks", which are copyrighted or private data. However, how can ChatGPT purchase copyrighted data and distribute it to its user is an unsolved problem. Obviously, traditional copyright laws are not applicable to the AI era. Commercialized ChatGPT and other AI tools cannot acquire data based on the old copyright laws.

Despite this, there are still ways for AI tools to access copyrighted data and improve their performance, while also benefiting data owners. Google's AdSense has given us some inspiration: a centralized platform created a project to connect to various websites, and distribute ads to these websites. The centralized platform gain revenue from advertisers through Google Ads, and share revenue with the websites that joined the project to distribute ads. This is one of the primary revenue streams for Google, and also one of the significant business models of the Web 2.0 era. The future AI era will be very different from Web 2.0 (see Section \ref{section_metrics} for more discussion), but the revenue-sharing model is still worth learning from. Whether ChatGPT's main revenue in the future comes from advertising, membership fees or other streams, it can share revenue with individuals and organizations that provide data to ChatGPT. This business model can also be adopted by other AI tools. In order to do that, we must design a new smart way to ensure that the centralized AI tools can share revenue fairly and transparently. 

Before proposing some specific methods and modeling for revenue sharing, let us discuss and emphasize why it is a good idea for AI tools such as ChatGPT to share revenue with its data providers.

\subsection{From AI war to AI-powered monopoly? Or revenue-shared utilitarian ecosystem?}\label{section_utili}

ChatGPT is popular, and perhaps Bard will catch up later, but we must recognize that ChatGPT and Bard are not the only LLM models available. 

Currently, the competition among LLMs and their chatbots is intensifying. In addition to OpenAI's GPT families, other LLMs include Google's PaLM \cite{PaLM}, LaMDA \cite{Lamda}, and \href{https://bard.google.com/}{Bard} which is powered by LaMDA, Meta's OPT \cite{OPT}, LLaMA \cite{LLaMA} and LLaMA-based simplified model Stanford's \href{https://crfm.stanford.edu/2023/03/13/alpaca.html}{Alpaca}, Amazon's AlexaTM \cite{AlexaTM}, Microsoft and NVIDIA's jointly launched Turing-NLG \cite{TNLG}, Baidu's Ernie 3.0 Titan \cite{Ernie} and recently launched \href{https://yiyan.baidu.com/}{ERNIE}, Anthropic's Claude \cite{Claude}. Moreover, there are also DeepMind's Chinchilla \cite{Chinchilla} and Gopher \cite{Gopher}, EleutherAI's GPT-NeoX \cite{GPT-NeoX}, BigScience's BLOOM \cite{BLOOM}, Bloomberg's BloombergGPT \cite{Bloomberg}, Huawei's PanGu-$\Sigma$ \cite{Huawei} and so on. All of these models are built upon the Transformer architecture or a derivative of it.

With various LLMs and their chatbots entering the AI market, there is likely to be fierce competition among different LLMs in the near future. While there are academic methods for evaluating the superiority of models \cite{HELM}, ultimately, the market will decide which ones will succeed. In my opinion, in order to win the LLM war, a successful LLM and its products must have the following characteristics: 

\begin{itemize}

\item \textbf{Having access to more and better quality data than other LLMs.} Just as discussed in Section \ref{section_more_data}, most LLMs are limited to training and finding answers only from the public domain data. Some LLMs, like BloombergGPT, utilize proprietary data, but this type of data is limited to what the companies control. If one LLM also use a more extensive knowledge from much more sources to provide better answers to a variety of questions beyond the public domain, that LLM will be definitely superior than others. 

\item \textbf{Having more powerful computing resources.} Massive amounts of training data required large deep learning models. As early as 2018, OpenAI found that the amount of computational power used to train the largest AI models had doubled every three to four months since 2012. Since then, the idea of LLM has been introduced and now it has become the mainstream of NLP. Currently, a couple of LLMs have hundreds of billions or even trillions of parameters. GPT-3, which has 175 billion parameters, took 3.64$\times 10^3$ GPU PF-days to train, with an estimated cost of 4.6 million dollars for a single training cycle\footnote{The cost estimate is from third party \href{https://lambdalabs.com/blog/demystifying-gpt-3}{OpenAI's GPT-3 Language Model: A Technical Overview}.}. If Huang's Law for GPUs will really replace Moore's Law for CPUs\footnote{Nvidia CEO Jensen Huang mentioned that in 2018 that the performance of GPUs is doubling every 18 to 24 months, which is faster than the rate predicted by Moore's Law for traditional CPUs.}, then we can expect that future models will become even larger with more complex architectures. Whoever can own more computing resources and have access to more data (including human feedback) will be able to produce better models. Given the high cost of computing resources, it is clear that only big technology companies or those with significant funding have advantage in LLM development.  
 
\item \textbf{Faster model iteration and refresh.} One of the biggest challenges for LLMs is how to update models. Transfer learning and fine-tuning may sometimes work, but it is better to refresh the entire models frequently. For example, the latest data ChatGPT used for training was from 2021. It will take long time for openAI to collect more recent data and upgrade the model. ChatGPT is almost ignorant of most recent things and may answer incorrectly, which contradicts the public's need to know the latests news and knowledge. The academic community has a disadvantage in updating models because they have limited data and computational sources than the industry, while big technology companies often update models more slowly than startups due to bureaucracy issues and a greater sense of moral responsibility, meanwhile startups usually lack computing resources and sufficient financial support. Perhaps a good approach would be startups and big companies to collaborate, and combine LLMs with search engines that have the latest information, just as Bard and Microsoft's New Bing are currently doing.  

\end{itemize}

%ChatGPT is rapidly gaining popularity, having acquired 100 million monthly active users just two months after its launch. OpenAI has announced that GPT will soon be updated to its fourth generation GPT-4, and as AI chatbots like ChatGPT continue to improve, they will become increasingly popular. Meanwhile, it is highly likely that similar competing products similar to ChatGPT will emerge in the market. The AI chatbot market will see various forms of competition, which will further promote the enhancement of AI chatbot performance, better meeting market demand, and people's usage and dependence of AI chatbots will gradually increase. 

As LLMs and AI chatbot technology become increasingly mature, perhaps people will increasingly rely on high-performing chatbots and gradually break free from their dependence on current search engines. People will use chatbots more frequently than they use Google search or any other search engines today. Just as Microsoft has integrated ChatGPT into its new Bing and 365 Copilot, Morgan Stanley has started using an OpenAI-powered chatbot, many LLM-based chatbot products have quickly entered the market, and begun to compete for user resources. In the foreseeable future, chatbots could generate enormous profits. Meanwhile, just as today's search engine market is dominated by Google search with over 90\% of the market share, one or several best LLMs with their chatbots could potentially dominate the chatbot or even multimodal AI market in the future. This is called \textbf{``LLM monopoly"} that the chatbot market will be dominated by only a few tools. In addition to chatbots, other AI domains may also see the emergence of monopolistic tools in the near future. Even multimodal AI tools may establish a monopoly that spans across multiple domains, without mentioning the sensitive topic of AGI (Artificial General Intelligence).

Let us call the future most popular chatbot as ``Chabot" for now and image its development. Chatbot will inevitably have a huge and growing training dataset. What would happen if Chatbot does not share revenue or profits with its data providers? Even if the data is currently in the public domain, it is highly likely that the original data owner will either lose more and more visitors and customers, or lose the credit of originality and creativity due to the impact of Chatbot. This will be a zero-sum game: the original data owner loses traffic, customers and reputation, while Chatbot uses the data to answer questions and earns revenue without given credit or quote anyone. Of course this is not acceptable. Even data that were once publicly available may start being copyrighted to prevent Chatbot from ``stealing" data, and to protect data owners' interests. As a result, Chatbot will only be allowed to use increasingly limited data due to copyright laws, resulting in lower quality or even constantly incorrect answers for customers. If there are still a large number of people relying on Chatbot at that time, the misleading responses from Chatbot may cause serious consequences. And if Chatbot loses popularity due to a decline in model performance, the AI revolution sparked by today's chatbots will suffer a serious setback, and we will very likely return to the era of Web 2.0. This is the consequence of the zero-sum game between AI tools and data providers. 

On the other hand, what if Chatbot were to share revenue with its data providers, regardless of whether their data is in the public domain or not? Even if the data provider loses customers and traditional reputation due to Chatbot, they can still benefit financially from the world's best chatbot. This is a virtuous circle: Chatbot collects more diverse, extensive and high-quality data to improve its performance, attract more customers, and gain greater revenue; while data providers can also get more revenue, and even get new reputation from sharing revenue with Chatbot. Once the shared revenue from Chatbot exceeds the loss caused by the loss of customers, it becomes economically beneficial for the data owner. We can be sure that even if Chatbot is willing to share revenue with data providers, there will still be ``old school" data owners who decline to share data. However, many other data owners will be encouraged to offer their data and participate in the Chatbot sharing revenue program.  It is truly a win-win plan for all parties, and a necessity for building a good AI ecosystem in the future. 

``Chatbot" is only an imaginary example for all popular AI tools in the future. In short, we may have two possible paths in the upcoming AI era. The first path is that AI tools have to use ``free data" only for training and profits from it. But even free data could be no longer free, if data owners consider AI tools as ``data thieves" and realize that it is a zero-sum game between them and AI tools, so the owners refuse to open their data any more. This creates a paradox for popular AI tools: on the one hand, they aim to become dominant players and potentially monopolies in the AI era, but on the other hand, the data they rely on is limited and even dwindling. If this is true, the AI era will be a hostile and toxic era. 

The second path shows a much better prospect which is what we hope to see: popular AI tools are willing to share revenue with their data providers, and more and more data owners are encouraged to participate in AI tools' revenue sharing programs as new data providers. As these AI tools obtain more diverse and high-quality data everywhere, they are able to constantly improve their performance, gains more users, earns more revenue, and shares more money with its data providers. This will be a utilitarian AI era where all parties benefit\footnote{According to Encyclopedia Britannica, \textbf{utilitarianism} is an effort to provide an answer to the practical question “What ought a person to do?” The answer is that a person ought to act so as to maximize happiness or pleasure and to minimize unhappiness or pain. This idea was proposed by 18th- and 19th-century English philosophers such as  Jeremy Bentham (1784-1832) and John Stuart Mill (1806–1873), but it is still applicable in the upcoming AI era. Although it is difficult to calculate "happiness", we should definitely build a good AI ecosystem and benefits as many ones as we can.}.  

Once we know that sharing revenue is a must for the AI era, we can then discuss the technical issues: how to make various AI tools share revenue? Let us examine how companies in the Web 2.0 era typically share revenue.
 
%Only in this way, can ChatGPT obtain more diverse, extensive, and high-quality data to improve its performance, gain more user favor, and achieve greater economic benefits. And only in this way, will various data providers be encouraged to offer their data and participate in this large-scale chatbot project that can benefit everyone in the world. It is truly a win-win plan for all parties, and a necessity for building a good AI ecosystem in the future. However, the traditional method of purchasing third-party data is too outdated and arbitrary for any AI tools, we must design a method to measure the importance of data, and establish a sound mechanism to distribute chatbot revenue with data providers. 

\section{Today's Revenue Sharing Models}\label{section_today_model}

Traditionally, revenue sharing refers to the distribution of a company's revenue among stakeholders, shareholders, and other contributors. In particular, some companies have adopted revenue sharing models to collaborate with public contributors such as photographers, artists, writers, and content creators for their role in the companies' success. Some companies create two-side platform to connect producers with consumers directly, while others create their own content and products with the assistance of public contributors, such as Getty images, Shutterstock, Adobe Stock, Google Adsense, Youtube, Amazon Associates and so on. Among them, Getty images and Goole Adsense provide two typical business models of sharing revenue with public contributors, and the current and future AI tools can also be inspired by their business models. 

\subsection{Getty Images}\label{section_getty}

Getty Images is a well-known American-British visual media company that specialized in stock images, editorial photography, videos and music for business and consumers. It has a large contributor base consisting of over 488,000 photographers and videographers as well as more than 300 content partners to deliver content to Getty Images for distribution \footnote{The numbers are from Getty images 2022 \href{https://investors.gettyimages.com/news-releases/news-release-details/getty-images-and-cc-neuberger-principal-holdings-ii-complete}{news report}.}. 

Getty images offers a revenue sharing model to its contributors. It sells stock photos and videos on-demand to consumers, then pays the original photo and video contributors royalty fees. For example, Getty images requests consumers a cost of \$375, \$1625 and \$3000  for a single, five and ten medium-size images respectively\footnote{More details see \href{https://www.gettyimages.com/plans-and-pricing}{Getty Image Pricing}.}, then pays 20\% as royalty rate to the image's provider. On the other hand, the royalty fee varies for iStock contributors. Exclusive contributors receive a default royalty of 25\% for stock photos, which increases to 30\%, 35\%, and 45\%, based on the total number of downloads.

Getty Images' revenue-sharing model encourages people to provide high-quality images and videos, thus ensuring that Getty Images has a vast stock images and videos. Based on the download volume of images and videos, the revenue-sharing rate has been divided into several fixed tiers. Unfortunately, this business model does not apply to AI tools. This because while Getty images contributors images and videos can be explicitly measured by download volume to determine their popularity,  the data providers for an AI tool cannot use any similar metrics to measure whose data is "more popular" with AI tool's customers. Although we can still assign some fixed revenue-sharing rates to different data providers, those rate will be arbitrary but not data-driven. Therefore, Getty images' revenue-sharing model cannot be applied to AI tools in the AI era. 

%However, the revenue-sharing rate only has a few fixed levels, which are determined by the download volume of images and videos. Such revenue-sharing model cannot be applied to AI tools in the AI era. If we make an analogy between the providers of training data for AI tools and the contributors to Getty Images, we find that it is impossible to directly link each set of training data for its outputs, in other words, the revenue obtained by the AI tool cannot be clearly allocated to each set of training data, and there is no ``download volume" metric for training data. 

\begin{figure*}[t]
\centering
\includegraphics[width=0.8\textwidth]{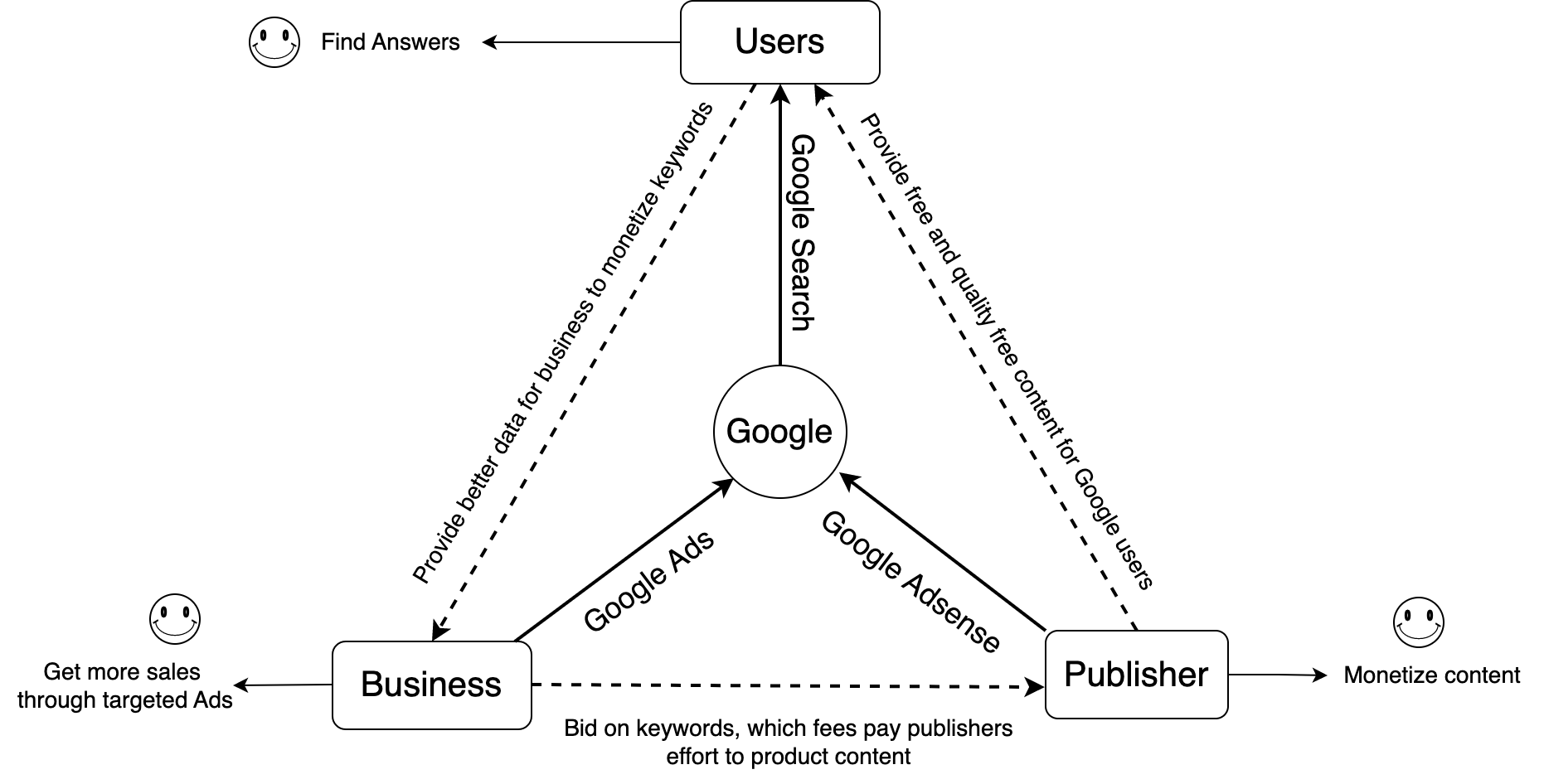}
\caption{Google's Business Model, adapted from \href{https://fourweekmba.com/google-business-model/}{Google Business Model Analysis}. Business join Google via Google Ads, while website publishers join Google by AdSense. Both ordinary users, business and publishers can benefit from this business model. However, this model cannot be used for AI tools  in the AI era.}\label{fig_google_model}
\end{figure*}

\subsection{Google Adsense}\label{section_adsense}

In the early days of the Internet, a majority of website owners had very limited options to earn revenue from their sites. Google introduced its \href{https://adsense.google.com/start/}{AdSense} program in 2003 and changed the game. Adsense is an online advertising platform that allows website publishers to display ads that were tailored to their content and audience, and Google took care of matching ads with website publishers, who can earn money when audience clicked on the ads. 

Figure \ref{fig_google_model} shows Google's business model. Google's advertising models, which includes \href{https://ads.google.com/}{Google Ads} (launched in 2000) and Google AdSense (launched in 2003), are among the most successful online business models in the age of Web 2.0, and the most important source of revenue for Google. Keep in mind that Google AdSense is different from Google Ads, although both of them are advertising platforms offered by Google. AdSense is for website publishers who want to display ads from Google and earn a portion of the revenue that Google earns from the advertisers who run the ads. On the other hand, Google Ads only run ads on Google centralized properties such as Google search results pages and YouTube. While content contributors to Google properties such as YouTubers can also benefit from Google Ads, Google AdSense has a much broader range of customers -- potentially any website publisher who joins the program can benefit from it. Which is better, Google AdSense or Ads? From the utilitarian perspective for Web 2.0, perhaps AdSense is better than Ads, although the former relies heavily on the latter. The latter is build on Google's centralized system, highly depending on Google, and the participants, whether they provide money or content, must contribute to Google (such as advertisers and YouTubers). The former is more like an ``alliance" established between a broad range of websites and Google. Websites do not need to directly contribute to Google, but can still share revenue from Google, making it a win-win system for both website publishers and Google. There are other programs similar to Google AdSense, such as \href{https://affiliate-program.amazon.com/}{Amazon Associates}. 

How much revenue does Google AdSense share with website publishers? Different from Getty Images, the revenue-sharing rate is not simply several fixed tiers, but depends on many factors. Advertisers pay Google on the basic of ads clicks, and website publishers receive 68\% of the ads revenue recognized by Google in connection with AdSense. For example, per click earnings depends on ads number of impressions, click through rate (CTR) and cost per click as
\begin{equation}
\textrm{Earnings} = \textrm{\# of impression}\times \textrm{CTR} \times \textrm{CPC} \nonumber
\end{equation}
where CTR and CPC vary depending on a variety of factors, including ads' categories and countries. Typically AdSense pays \$0.2 - \$2.5 per 1,000 views on average. Although the entire revenue sharing system is far more complex than Getty Images, participants in Google AdSense can clearly see their earnings on their platform, and the overall process is still transparent. Other programs may have different way to pay, but the principle of revenue sharing is more or less the same. For example, Amazon Associates allow participants to earn a commission when someone clicks on their unique affiliate link and makes a qualifying purchase on Amazon. The commission rate varies depending on the product category and ranges from 1\% to 10\% of the sale price.

%pay publishers typically \$8-20 for 1000 views on average. The earnings not only depend on the website traffic, but also depend on web category and content, as well as web locations.  Per click revenue is calculated as click-through rate * cost per click * number of impressions/100. 

In the upcoming AI era, there may be new forms of advertising, or we can assume that a large source of revenue for various AI tools will come from membership fees of premium customers. Let us consider whether the principle of AdSense can be borrowed by AI tools. While there is no click and click-through-rate in the entire process of customers inputting prompts and AI tool generating responses, we can still analogize every single prompt input as a click. For AI tools, \textbf{Cost Per Action (CPA)} in the Web 2.0 era can be replaced by \textbf{Cost Per Prompt (CPP)}. Although the customers of AI tools do not pay based on the number of prompt inputs, we can still establish a scoring system for AI data providers by calculating the connection between each prompt input and all data providers, and then share revenue among data providers based on the scoring system. In the next chapter, we will examine how to establish such a CPP scoring system for data providers in detail. 

%and we have argued in Section \ref{shareRev_good} that AI tools' revenue must be shared with their data providers. In this regard, the business model of sharing revenue with AI tools is somewhat similar to the business model of Google AdSense. Of course, the differences between the two models are also quite obvious: in the AI era, we cannot use CPA to calculate the usage of various training datasets behind AI tools. We must use more intelligent methods to measure the  usage of various data, so that those training datasets can be reasonably distributed the revenue of AI tools. 

%A publisher will get 68\% of the click amount and the commission per click ranging from \$0.20 to \$15.  
%Revenue sharing model works pretty well for Web 2.0 online ads business based on CPA , but it does not  does not work for AI models.  
 
\begin{figure*}[t] 
  \centering
  \includegraphics[width=0.9\textwidth]{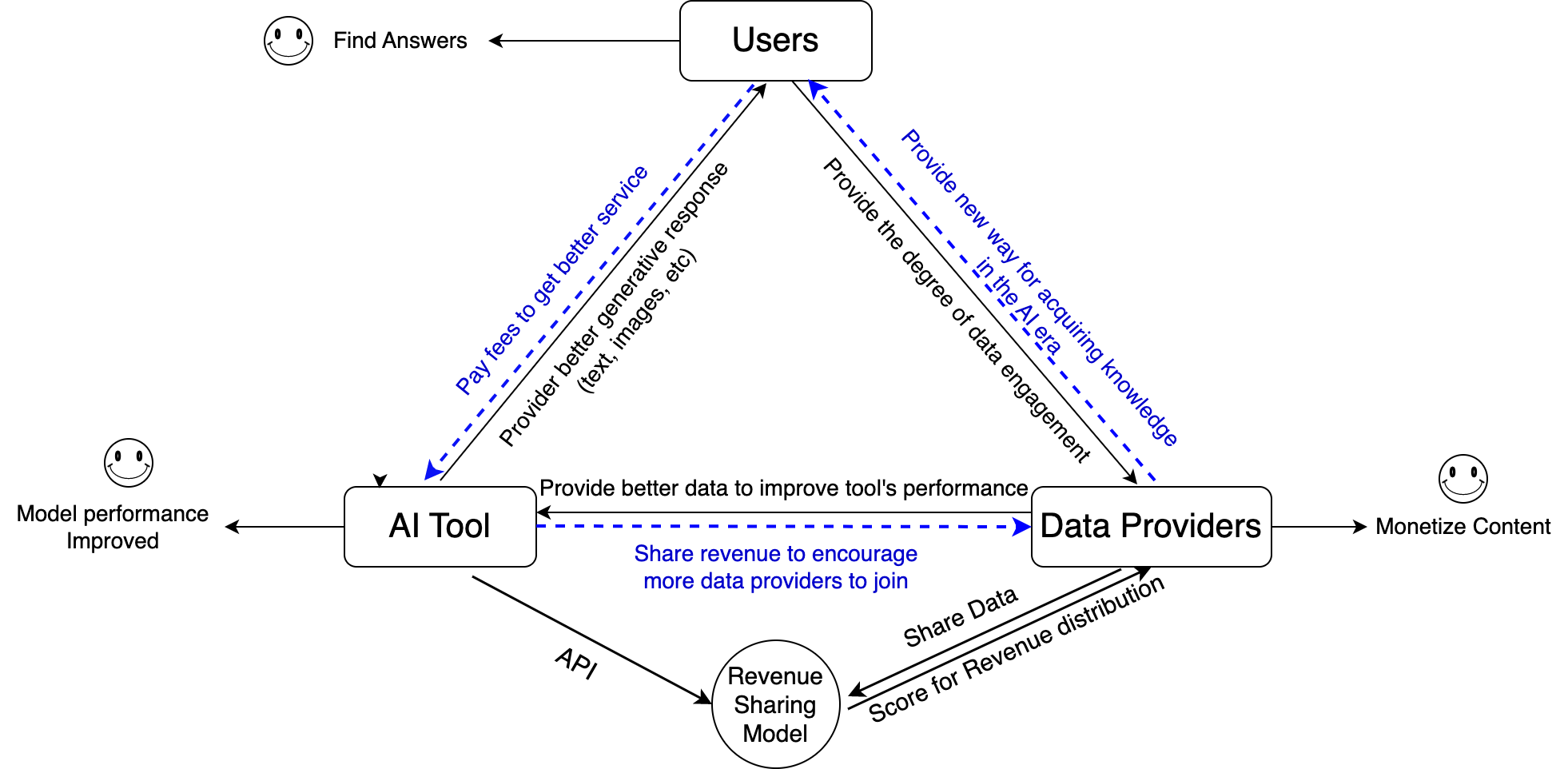}
  \caption{Proposed business model for AI tools in the AI era. It includes four key elements: the AI tool itself, the users, the providers of data for the AI tool, and the model for revenue sharing. If there is a continuous inflow of data providers into the AI tool's ecosystem, similar to Google's business model (compare to Figure \ref{fig_google_model}), users can receive better generative responses due to AI model improvement, and data providers can earn revenue. However, there are significant differences between this business model and Google's model. For instance, users may need to pay fees to access better services instead of using them for free. Also, data providers cannot directly show their content to users, the content is transformed by the AI tool and displayed to users after AI tool's model training. Additionally, the revenue-sharing model is independent of the AI tool itself.  See Section \ref{section_scoring_require} for more discussion.}
  \label{fig_share_rev_model}
\end{figure*}

\section{A New Revenue Sharing Model for AI Data Providers}\label{revenue_share_model_AI}

Figure \ref{fig_share_rev_model} demonstrates the new revenue-sharing business model, including its connection to the AI tool, the data providers for the AI tool, and the tool's users. Similar to current Google's business model, all parties (users, data provides and AI tool) can benefit from this model. However, the difference between this model and Google's model is obvious: most metrics used for Web 2.0 business model need to be replaced by new metrics. Meanwhile, there is a separated revenue-sharing model, which is crucial to calculate how to distribute revenue to the AI tool's all data providers.

\subsection{Old Metrics and New Metrics}\label{section_metrics}

In the Web 2.0 era, people used to ask the question: ``how to measure online ads' importance and success?" In the AI era, the corresponding question is not how to measure the importance of an AI tool, but how to measure the importance of each data provider for an AI tool. We can see that the metrics used to measure ads importance are essentially ineffective for AI tools:

\begin{itemize}

\item[(1)] \textbf{Page views}. AI tools may still be integrated into various websites, but the count of web page views no longer matches the usage of AI tools. For example, the web-based ChatGPT and Bard can be used for a long time without refreshing the page. The relevant ``bound rate" and ``average view duration" will be less important for AI tools.

\item[(2)] \textbf{Clicks}. The click that originated from hypertext markup language (HTML) may no longer be applicable in the AI era, as AI tools no longer reply on clicks to function. Consequently, many important metrics associated with clicks, such as Click Through Rate and Cost Per Click,  are no longer suitable for weighing the importance of AI products or the data provided for AI training.

\item[(3)] \textbf{Conversion rate}. The current conversion rate of an advertisement is closely related to clicks and purchases. However, for an AI tool, although it is still possible to calculate the proportion of free users and premium users, such calculations do not help in understanding how the AI training data is used and being utilized. 

\item[(4)] \textbf{Engagement rate}. Similar to the conversion rate, the old method of calculating the user engagement rate is no longer applicable for evaluating the data importance in AI tools. 

\end{itemize}

So what are the most important metrics for an AI tool?  These metrics as follows:

\begin{itemize}

\item[(I)] \textbf{Prompts}. It is highly likely that prompts will replace clicks as the most important metric in the AI era. Each input prompt from a user contains certain information, triggering AI tools to response. The traffic and information of prompts will become the most important measurement of an AI tool in the visible future. 

\item[(II)] \textbf{Cost Per Prompt}. As mentioned at the end of the previous Section \ref{section_adsense}, Cost Per Click will be replaced by Cost Per Prompt. How does an AI tool call its pre-trained or fine-tuned model to generate a response for each prompt, and how are prompts and generated responses related to the AI tool's training dataset? These are crucial questions. 

\item[(III)] \textbf{Data Engagement Rate}. Let us qualitatively brainstorm the concept of ``data engagement". Consider an AI chatbot that has been trained on a large amount of text data about physics and astronomy. This chatbot is likely to be able to answer many questions on physics and astronomy, but it may not know anything about Shakespeare's plays. Therefore, if a prompt is related to physics or astronomy, we can say that the training data of this chatbot is engaged, or has a high \textbf{``degree of engagement"}. On the other hand, if a prompt is to ask about Shakespeare, we say that the chatbot's data is not engaged, or has a very low degree of engagement for this prompt. If an AI tool has may classes of training data across a wide range of diverse topics and areas, the degree of engagement of each class is obviously different for an input prompt. By quantifying and normalizing the engagement degree, we can calculate the data engagement rate. 

\end{itemize}

\subsection{Scoring System and Requirements}\label{section_scoring_require}

In order to calculate the new metrics mentioned above such as Data Engagement Rate for AI tools, we must build a scoring system, where each data provider for an AI tool can have a score. Using this scoring system, we can calculate the Cost Per Prompt for each data provider and the degree of engagement of each data provider for prompts. Then, we can use the engagement scoring system to calculate the AI tool's share of revenue for each data provider.  

To build a scoring system, there are several crucial prerequisites:

\begin{itemize} 

\item \textbf{The scoring system is based on prompts.} As mentioned in Section \ref{section_metrics}, for AI tools, the most important metric is prompts rather than clicks. Therefore, the scoring system should be based on each prompt, not each click. Specifically, when a user inputs a prompt for an AI tool, each data provider of the tool will receive a score for that prompt. These scores for all prompts assigned to data providers are linearly summed up to generate a final score for providers, which is used to measure the engagement degree of the data they provided. 

\item \textbf{The scoring system should be as simple as possible} -- DONOT use deep neural network to build it. Nowadays, there are more and more AI tools, a considerable number of which employ deep learning models. Various LLMs and large computer vision models have reached the scale of tens and hundreds of billions, even trillions of parameters, with complex deep neural network architectures. Such large deep learning models are often business secrets and not open to the public, and more importantly, difficult for the public to understand. The scoring system is for general data providers, we must ensure that the scoring system can be explained to and understood by the public, which means the scoring system as a model itself, should be as simple as possible. For example, the scoring system can be based on a tree-based model or vector calculation rather than relying on the AI model neural network architectures.

\item \textbf{The scoring system is build for all data providers of an AI tool, we need to ensure that the AI tool's entire training dataset is open and transparent, allowing third party verification.} Some large deep learning models have publicly released their training datasets, but there are also many tools and their models' training datasets are not publicly available. There are various reason for this, such as some datasets include proprietary content, meaning that the companies that developed the models have exclusive rights to use them. Some datasets may contain personal information or other sensitive data that could be used to identify individuals or compromise their privacy. Moreover, legal restrictions or concerns about data bias may also cause companies and organizations to hesitate in opening the entire training datasets of their models. To establish a scoring system that applies to all data providers while avoiding the issues mentioned above, the following solution can be employed: the dataset used by an AI tool is not made public, but only shared with a reliable third party that is responsible for building the scoring system. Brining in a third party ensures a certain level of transparency for the AI model's training dataset, while also avoiding the dataset being made open to the public.

\item \textbf{The scoring system must be as independent from the AI tool as possible.} Whether the scoring system is built by a third party or by the company or organization that created the tool, we do not want the scoring system to be too closely related to the tool's model itself. This is because the more they are related, the more the scoring system depends on the model itself, the harder to explain the scoring system to the public. We need to ensure that even if we treat the AI model as a black box, the scoring system can still be established and explainable.  However, when I say ``as independent as possible", it means that the scoring system and the AI tool can also have some connection, but the connection can only be at the API level. That is to say, the AI tool itself provides API-level information to the scoring system, such as doing document embedding through API (will be discussed more detailed in later sections), and the scoring system builds independently based on this information.

\end{itemize}

Once the scoring system is established, we can use it to evaluate the engagement of each data provider and distribute revenue accordingly. Let us now take a look at the possible mechanisms for building the scoring system. 

\subsection{Classification}\label{section_class}

The most straightforward approach to create a scoring system is to build a supervised classification model, which can generate probability scores. For the training dataset of an AI tool, if we consider each data provider as an independent class, or a combination of data providers as a class, so each document belongs to one class, and we can build a pre-trained classification model on this training dataset. Once a user inputs a prompt and receives a generative response from the tool, we can use the pre-trained classifier to assign each class a probability score based on the prompt or the response, so the probability scores reflect the degree of engagement of each class for this prompt. The summation of probability scores for all prompts can be used to establish a scoring system.

%%Based on the classification model, each document can be assigned a series of probability scores, which can be used to establish a scoring system. % The most straightforward approach to create a scoring system is to develop supervised classification models, which provide probability score to each classes. Consider that the training set of an AI tool comes from various providers, and each provider can be treated as one independent class. Therefore, each document used for training can be labeled for one class, then a text classifier can be built on all training documents. 

%build above to measure the ``degree of engagement" of each class in the training dataset for the input and response texts. Since the text classifier can provide the probabilities of the input and response text belonging to each class, the probabilities can be used as score to measure the engagement of each class, i.e., each provider for this case. 

\vspace{0.06in}

\subsubsection{\textbf{Newsgroup20 Demonstration}}

To better illustrate how it works, we can use the benchmark dataset \href{https://scikit-learn.org/0.19/datasets/twenty_newsgroups.html}{Newsgroup20} to demonstrate how to build a classifier on a series of text documents and calculate the probabilities that any text document belongs to each class. \textbf{Newsgroup20}, which was originally collected in 1995-1996 from the ``Newsgroups" project, is a widely used dataset for text classification tasks, consisting of around 20,000 documents distributed among 20 distinct newsgroups. The dataset has been split into training and testing sets, where training set contains $\sim$11.3k documents and testing set contains $\sim$7.5k documents. We can view each newsgroup as a data provider -- imagining that Newsgroup20 is composed of data provided by 20 different data providers. If we treat each provider as a class, we can use the training set to built a text classifier for these 20 classes, and calculate the probabilities of any text document belonging to each class. 

The first step is to do text embedding and vectorize documents. Text embedding is more sensitive for calculating text similarity, which will be discussed in Section \ref{section_text_sim}.  Here, as a baseline model,  we utilize the widely-used TF-IDF technique, which includes \texttt{\small CountVectorize} and \texttt{\small  TfidfTransformer} to tokenize and vectorize each text document, and use \texttt{\small  LinearSVC} and \texttt{\small CalibratedClassifierCV} to train and build the text classifier.  Next, we can apply the classifier to any text documents. Let us randomly pick up a document in the testing dataset. For example, there is a document which talks about Hubble Space Telescope (HST, hereafter Document HST) :

\begin{lstlisting}
From: henry@zoo.toronto.edu (Henry Spencer)
Subject: Re: HST Servicing Mission Scheduled for 11 Days
Organization: U of Toronto Zoology
Lines: 12

In article 1993Apr27.094238.7682@samba.oit.unc.edu 
Bruce.Scott@launchpad.unc.edu (Bruce Scott) writes:

If re-boosting the HST by carrying it with a shuttle would 
not damage it, then why couldn't HST be brought back to 
earth and the repair job done here?

The forces and accelerations involved in doing a little bit 
of orbital maneuvering with HST aboard are much smaller 
than those involved in reentry, landing, and re-launch. 
The OMS engines aren't very powerful; they don't have to be.

SVR4 resembles a high-speed collision | Henry Spencer 
@ U of Toronto Zoology between SVR3 and SunOS. 
- Dick Dunn | henry@zoo.toronto.edu utzoo!henry
\end{lstlisting}

%\begin{lstlisting}
%From: perky@acs.bu.edu (Melissa Sherrin)
%Subject: Re: moon image in weather sat image
%Organization: Boston University, Boston, MA, USA
%Lines: 14
%Originator: perky@acs.bu.edu

%I'm afraid I was not able to find the GIFs... is the list 
%updated weekly, perhaps, or am I just missing something?

 %  _______
 % (       )
% (_  (     )
%   (      )
%  (    )  )
%(  (     )
%(__________)
 %/ / / / /
% Melissa Sherrin
% perky@acs.bu.edu
%\end{lstlisting}

\begin{table}[htbp]
\centering
 \setlength\arrayrulewidth{0.5pt}
\renewcommand{\arraystretch}{1.2}% Tighter
\label{table2}
\begin{tabular}{l|lll}
\hline
\textbf{Newsgroup20} & \textbf{Score 1} & \textbf{Score 2} & \textbf{Score 3} \\ \hline
sci.space & 0.9527 & 322.2 &  0.6710 \\  
rec.autos & 9.05e-3 & 2.859 & 1.77e-3 \\ 
sci.electronics & 5.16e-3 & 10.10 &  1.67e-3 \\ 
comp.sys.mac.hardware & 4.77e-3 & 4.48 & 0.200 \\
misc.forsale & 4.06e-3 & 3.91 & 6.07e-3 \\ 
soc.religion.christian & 3.29e-3 & 2.78 & 4.88e-3 \\ 
rec.sport.hockey & 2.92e-3 & 1.22 & 8.50e-3 \\ 
rec.sport.baseball & 2.87e-3 & 1.26  & 3.30e-2 \\ 
comp.graphics & 2.43e-3 & 10.2  & 2.13e-3 \\ 
alt.atheism & 2.17e-3 & 2.51 & 6.56e-4 \\ 
comp.sys.ibm.pc.hardware & 2.10e-3 & 4.08 & 1.25e-2 \\
comp.windows.x & 1.56e-3 & 4.10 & 1.84e-2 \\ 
comp.os.ms-windows.misc & 1.54e-3 & 2.34 & 6.11e-3 \\ 
talk.religion.misc & 1.50e-3 & 2.18 & 1.46e-2 \\ 
talk.politics.misc & 1.30e-3 & 4.00 & 3.41e-3 \\ 
sci.med & 1.10e-3 & 7.81 & 2.95e-3 \\ 
talk.politics.guns & 5.37e-4 & 3.01 & 1.48e-3 \\ 
talk.politics.mideast & 4.45e-4 & 1.04 & 4.78e-3 \\ 
rec.motorcycles & 2.72e-4 & 1.90 & 5.33e-3 \\
sci.crypt & 1.90e-4 & 1.99 & 3.34e-4 \\ \hline
\end{tabular}\caption{Distribution of Newsgroup20 probabilities given by the text classifier. Score 1 are the probability scores of one document (Document HST) from Newsgroup testing dataset, Score 2 are the summed up probability scores of all document labeled as ``sci.space" from Newsgroup testing dataset, while Score 3 are the score of a given prompt input "Why the sky is blue?".}
\label{tab_newsgroup20}
\end{table}

We use the text classifier to show the probabilities of the Document HST belongs to each class. Score 1 in Table \ref{tab_newsgroup20} gives the result: this document has 0.9527 probability of belonging to the class ``sci.space", which is consistent with its real labeled class. On the other hand, the document has lower probabilities of belonging to other classes, such as 0.009 for ``rec.auto", and 0.005 for ``sci.electronics". We can consider the twenty probability scores as scores for a single document belonging to each of the twenty classes respectively. Just for the Document HST, ``sci.space" has the highest engagement due to its highest probability score. 

The probability scores can be summed up for many documents. For example, if we collect all the documents labeled as ``sci.space" in the Newsgroup20 testing dataset, which totally has 394 such documents, the summed up score for ``sci.space" is 322.2, and for ``comp.graphics" is 10.2. Score 2 in Table \ref{tab_newsgroup20} provides the score corresponding to each class. Note that the sum of all these scores is 394, which is the number of documents. If we consider that each document weights equally, we can normalize Score 2 by dividing it by 394 to obtain the normalized probability scores. Again, the class ``sci.space" has the highest degree of data engagement, followed by ``sci.electronics" and ``comp.graphics". It is easy to understand why the class ``sci.space" has the highest degree of engagement, since all documents used in this case are from this class. However, for other classes, the level of engagement may not be as apparent or clear. For instance, why does the degree of engagement of ``sci.electronics" exceed that of ``rec.autos"? To answer such questions, Explainable AI (XAI) is a commonly used method to explain machine learning decisions such as text classification and probability score calculation. In particular, SHAP can be used for NLP model explainability \cite{NLP_survey, SHAP_NLP_CNN}. However, XAI is beyond the scope of this paper, we expect to discuss it in future work. 

%The text classifier can classify and score any input content, including documents, paragraphs and sentences. 

Next, let us take a look at the probability scores assigned by the aforementioned text classifier to any given text input. Score 3 in Table \ref{tab_newsgroup20}  gives the score distribution for a given input prompt: {\ttfamily{\small Why the sky is blue?}} We can see that the highest probability is still from ``sci.space", which is understandable, followed by the probability of ``comp.sys.mac.hardware", which is less obviously. As mentioned before, we may need XAI to better understand the probability scores, but at least the text classifier indeed give a prompt-based scoring system to measure the degree of engagement of each class for any documents.  

\begin{figure}[t] 
  \centering
  \includegraphics[width=\linewidth]{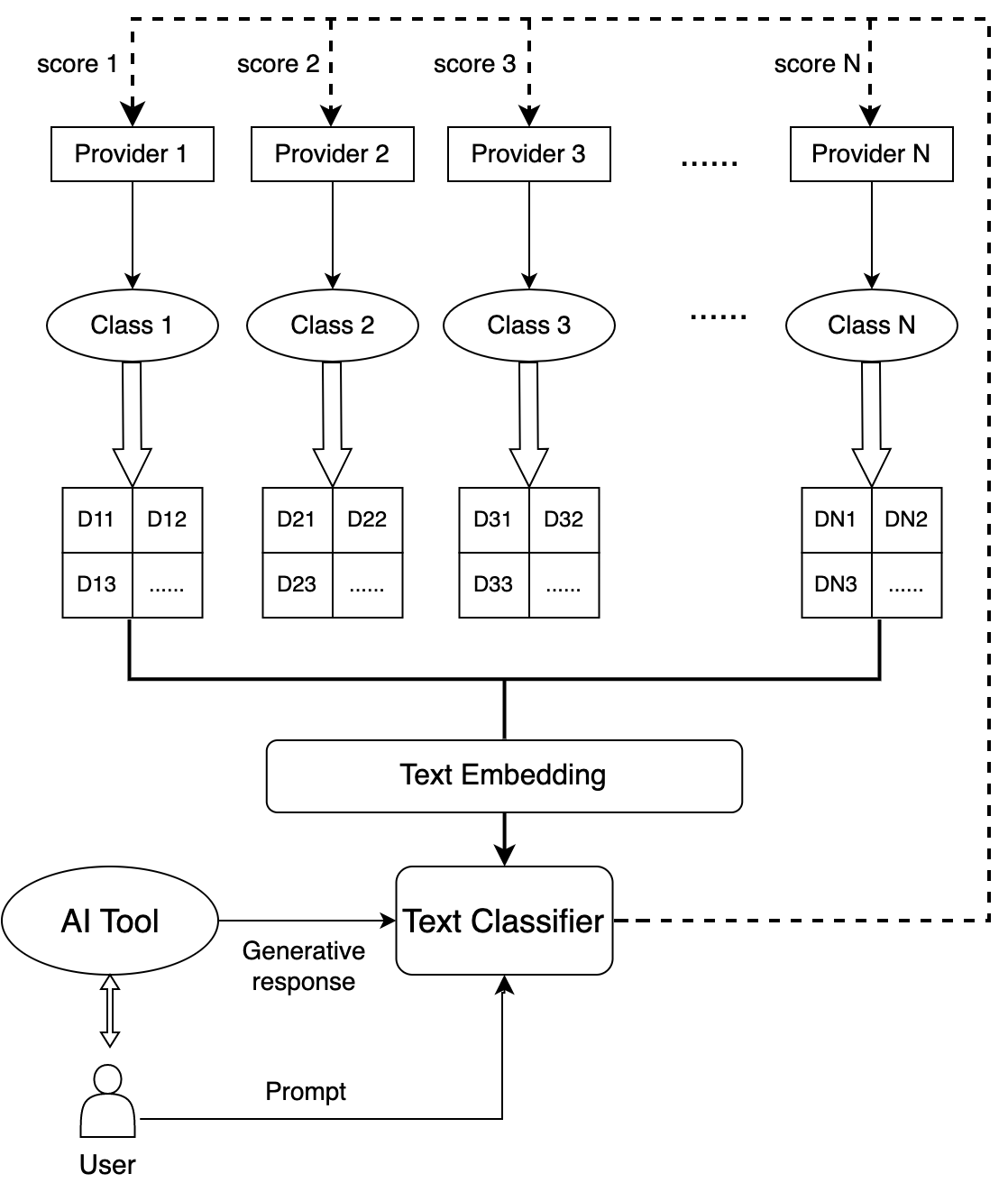}
  \caption{The revenue-sharing scoring system based on text classification. Each data provider can be considered as one class (or a cluster of data providers as on e class), with a series of documents in each class (D$_{i1}$, D$_{i2}$, D$_{i3}$ for $i-$class). Documents are vectorized using some text embedding techniques (see Section \ref{section_text_sim} and Figure \ref{fig_similarity} for more discussion on text embeddings), then a text classifier can be built for all documents. A user inputs a prompt in the AI tool and receives generative response, both the prompt and response can go through the text classifier to generate a series of probability scores for each data providers. }
  \label{fig_text_classifier}
\end{figure}

\vspace{0.06in}

\subsubsection{\textbf{Extend to General Case}}

In principle, we can apply this idea to any AI tools as well. Figure \ref{fig_text_classifier} shows the basic pipeline for establishing a scoring system based on a text classifier. Each data provider, or a group of data providers can be treated as a class. With each document assigned a class, we can use AI tool's all training data to train a text classifier, which can be used to establish a score system for any text documents. From the perspective of revenue sharing, if each class represents a data provider, who are able to share revenue from the AI tool with other providers, then the revenue that one provider can share is calculated as: 
\begin{equation}
\mathscr{R}_{i} = \mathscr{R}_{\rm tot} \times P_i, \label{equ_RI}
\end{equation}
where $\mathscr{R}_{\rm tot}$ is the total shared reveue, $\mathscr{R}_{i}$ is the revenue shared by $i-$th provider, and $P_{i}$ is the provider's normalized score calculated by
\begin{equation}
P_i = \sum_{n}p_{in}/\sum_{i,n}p_{in}
\end{equation} 
and $p_{in}$ is the probability of the $n-$prompt input or its response belonging to the $i-$th provider's dataset. 

%In summary, we can build a text classifier on the training dataset, with each document assigned a class,  and use the classifier to establish a score system for any text documents (including paragraphs and sentences). 

%\textbf{Gained Revenue by One Provider $=$ Total Shared Revenue from the AI tool $\times$ Normalized Probability Score for this Provider}.

%In principle, we can apply this idea to any AI tools as well. 

\vspace{0.06in}

\subsubsection{\textbf{GPT as Another Case} }

The full training datasets for ChatGPT and GPT-4 have not opened to the public. Let us look into GPT-3. We can consider the training dataset of GPT-3 comes from the following five providers: Common Crawl, WebText2, Books1, Books2, and Wikipedia, and treat each provider as one class to build a text classifier on the training dataset. A somewhat challenging question is how to decide on the size of each text document\footnote{This is not a problem for image or video dataset, since each image or video can be treated as one document.}. This is also a general question not only for GPT, but also for all text datasets. For example, if one data provider offers three documents, each of which has millions of tokens in length, while another data provider offers hundreds of documents, each with only one sentence, it is clear that the sizes of their documents are very different. One approach is to consider using truncated documents to roughly equalize their length. For GPT models themselves, each input document is typically a sequence of contiguous text tokens. The original model used a fixed-length input sequence of 1024 tokens, while GPT-3 uses a more flexible input format. For text classification, we can truncate text to make a single document to have a size that is roughly equal to the average length of generative responses.

For any prompt input by a customer, the pre-trained text classifier can assign scores to each of the aforementioned five classes. The summation of these scores can indicate the degree of engagement of the five data providers in all generative responses, thus determining how much revenue each provider can share -- if ChatGPT would like to share revenue with them.  Of course, dividing GPT-3 into five data providers is just for brainstorming. In the future, if we have hundreds or thousands or even more data owners providing training data for GPT, we can still label these large number of data providers as different classes to build a text classifier, and calculate each provider's degree of data engagement by classifier probability scores. 

\vspace{0.06in}

\subsubsection{\textbf{Large Number of Data Providers}}

\begin{table}[htbp]
\centering
 \setlength\arrayrulewidth{0.5pt}
\renewcommand{\arraystretch}{1.2}% Tighter
\label{table2}
\begin{tabular}{l|lll}
\hline
\textbf{Reuters-21578} & \textbf{Score 1} & \textbf{Score 2}  \\ \hline
earn & 780 & --  \\
acq & 7.84 & 0.538 \\
money-fx & 3.71 & 1.52e-2 \\
ship & 1.63 & 1.48e-2 \\
grain-corn & 1.62 & -- \\
crude & -- & 2.37e-2 \\
interest & -- & 1.21e-2 \\
 \hline
\end{tabular}\caption{Top five probability scores for all documents in the Reuters-21578 testing dataset labeled as ``earn" (Score 1), and for the prompt int  ``Why the sky is blue?" (Score 2).}
\label{tab_reuters}
\end{table}

If an AI tool has many data providers, and each provider is viewed as a separate class, then building a text classifier using the training dataset of this tool can be considered as a multi-class problem. We can use another benchmark dataset, \href{https://archive.ics.uci.edu/ml/datasets/reuters-21578+text+categorization+collection}{Reuters-21578}, to demonstrate for many-class problem. The Reuters-21578 dataset consists more than 21K newswire articles from the Reuters news agency, which were collected in 1987 and used for text classification research. The articles are classified into 90 different topics, but we can further divide the classes to create more specific ones. For example, if a document belongs to both the ``grain" and ``wheat" classes, we can assign it to a new class called ``grain-wheat". Similarly, if a document belongs to the ``interest", ``retail" and ``ipi" classes, we can create a new class called ``interest-retail-ipi". As a result, the Reuters-21578 training set is divided into a total of 465 classes. We can build a text classifier on the training dataset for these classes using the same method mentioned above for the Newsgroup20 example. 

Next we test the text classifier probability scores by taking all 1041 documents in the Reuters-21578 testing dataset that are labeled as ``earn", seeing how the classifier assigns probability scores to these documents. Score 1 in Table \ref{tab_reuters}  gives the result for five classes with highest scores. The summed up probability score of ``earn" for these documents is 780, which is the highest score, followed by significantly lower scores from other classes. When we test another single prompt {\ttfamily{\small Why the sky is blue?}} , the text classifier assigns the highest probability score to the class ``acq", followed by ``money-fx", ``crude", ``interest"  and so on. It may be difficult to understand why Reuters-21578 would classify a question about the sky as being highly related to corporate acquisitions (acq). One main reason is that the training dataset of Reuters-21578 is incomplete, which means the training data does not cover enough topics, for example, topic related to ``space". When there is a lack of training data for a specific topic, the classifier is incomplete, and can only assign the input text to the most similar class. If this were to happen in a real AI tool, we do not expect the AI tool with an incomplete training dataset has a suitable response for a prompt which topic is not covered in that dataset.

%is incomplete. Unlike Newsgroup20, Reuters-21578 has very few topics related to ``space". When there is a lock of training data for a specific topic, the classifier can only assign the input text to the most similar class. For more details, we need to use explainable Natural Language Processing (NLP) to explain the result. %and the scores for other classes are significantly lower. This is understandable. However, when we test another single prompt input {\ttfamily{\small Why the sky is blue?}} , the text classifier assigns the highest probability score to the class ``acq", followed by ``money-fx", ``crude", ``interest"  and so on. It may be difficult to understand why Reuters-21578 would classify a question about the sky as being highly related to corporate acquisitions (acq). One reason is that the data in the training set is incomplete. Unlike Newsgroup20, Reuters-21578 has very few topics related to ``space". When there is a lock of training data for a specific topic, the classifier can only assign the input text to the most similar class. For more details, we need to use explainable Natural Language Processing (NLP) to explain the result. 

Therefore, building a scoring system based on probability scores of a text classifier on the training dataset of an AI tool can be comprehensible, but its interpretation could be challenging if the training data is incomplete or poorly labeled. In such scenarios, an alternative scoring method based on unsupervised techniques like text similarity can be developed for any text.

\begin{figure*}[t] 
  \centering
  \includegraphics[width=1\textwidth]{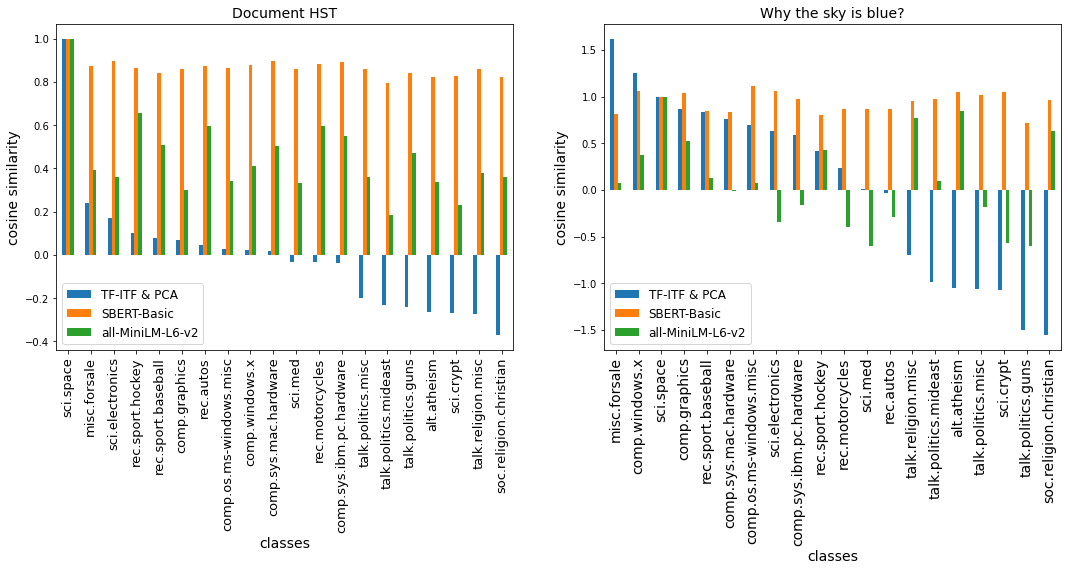}
  \caption{Cosine similarity between Newsgroup 20 classes and the Document HST (left), and the prompt {\ttfamily{\small Why the sky is blue?}}. Multiple text embedding techniques were used to vectorize text documents: directly TF-IDF and PCA to reduce vector dimensions, pre-trained models including SBERT-Basic and all-MiniLM-L6-v2. }
  \label{fig_similarity}
\end{figure*}

\begin{figure*}[t] 
  \centering
  \includegraphics[width=\linewidth]{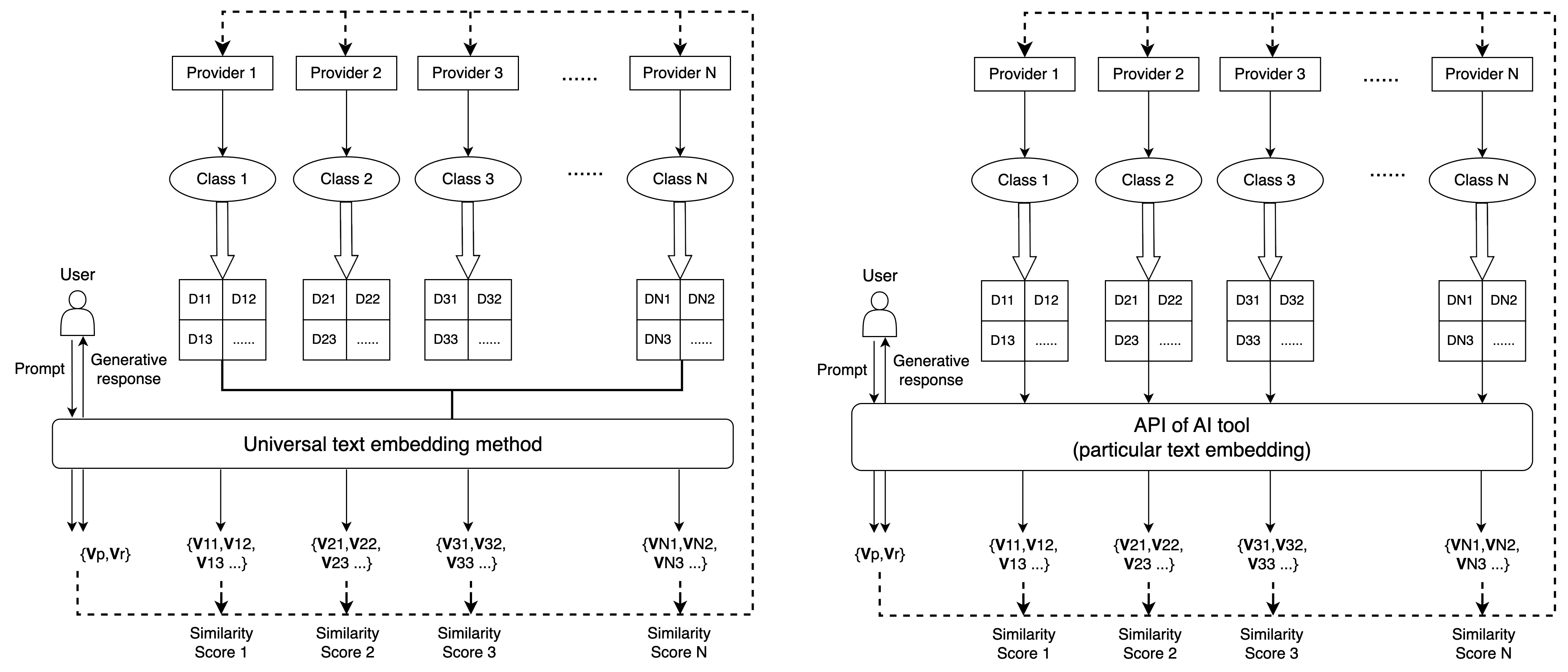}
  \caption{The scoring system based on text similarity. Documents can be vectorized using two types of text embedding techniques: universal text embedding method (left panel) and tool/model-based particular text embedding method (right panel). A user's prompt and the generative response from the AI tool can be vectorized using the same methods, and calculate the average similarity between documents in each class and prompt/response. The average similarity scores are used to establish the scoring system.}\label{fig_sim_score}
\end{figure*}

\subsection{Text Similarity}\label{section_text_sim}

Tex similarity is an unsupervised technique to measure how similar two or more pieces of text are to each other in terms of their meaning or content. For example, let us consider the similarity among the following three sentences: 
\begin{itemize}

\item {\ttfamily{\small Why the sky is blue?}} (Sentence I)
\item {\ttfamily{\small Why the space is dark?}} (Sentence II)
\item {\ttfamily{\small The sky is blue due to a phenomenon called Raleigh scattering.}} (Sentence III, answer to Sentence I). 

\end{itemize}

The commonly used method to compare text similarity is to use some text embedding techniques to vectorize all texts, and measure the distance between vectors in a high-dimensional space. The measurement techniques including \textbf{cosine similarity}, \textbf{word embedding-based similarity}, and \textbf{latent semantic analysis}. Using TF-IDF method to vectorize and cosine similarity to measure the distance for the above three sentences, we have the similarity matrix:
\begin{equation*}
\begin{bmatrix}
1 & 0.452 & 0.424 \\
0.452 & 1 & 0.138 \\
0.424 & 0.138 & 1
\end{bmatrix}
\end{equation*}
which means the first sentence is more similar to the second sentence than the third sentence, while the second and third sentences have the lowest similarity. 

\vspace{0.06in}

\subsubsection{\textbf{Universal Text Embedding Method and Cosine Similarity}}

Next, let us use Newsgroup20 again to demonstrate how to establish similarity between any documents and the various classes in a dataset. The first thing is always to find a text embedding technique. Images and videos may need some other techniques (see Section \ref{section_text2image}), let us focus on text embeddings in this section. 

We still use Newsgroup20 to demonstrate how to embed and vectorize text documents. In principle, the TF-IDF method provides a technique to perform text embedding for documents in the Newsgroup20 training dataset and convert documents to vectors. As a result, each training document in Newsgroup20 is vectorized into a 130093-dimensional vector. And we can vectorize any document into the same dimensional vector, and calculate its cosine similarity between vectors to show the degree of similarity between any documents. However, since the vectors generated by TF-IDF are of high dimensions and are also very sparse, such calculations are usually time and energy consuming. To simplify the calculations, a commonly used method is to reduce the dimensions. As an example, I used Principle Component Analysis (PCA) to reduce each vector in the training set from 130093 dimensions to 768\footnote{The number 768 is the dimensions of text embedding from base-Bert. I use the same number of dimensions for a comparison with BERT. }, and used the same method to reduce any vectorized document to 768 dimensions.   

Let us look at the Document HST mentioned in Section \ref{section_class} about the Hubble Telescope. After text embedding, we calculated the cosine similarity between each document in the training set and the Document HST. The summation of cosine similarity between the Document HST and the $i-$th class can be calculated by
\begin{equation}
S_{\textrm{HST},i} =   \sum_{j=1}^{K} \textbf{V}_{\rm HST}  \cdot \textbf{V}_{ij}/|\textbf{V}_{ij}|\label{equ_HST}
\end{equation}
where $\textbf{V}_{\rm HST}$ is the embedded Document HST as a vector, $\{\textbf{V}_{ij}\}$ is embedded documents in the $i-$th class with $j=1,2,3 ... K$ and $K$ being the number of documents in that class. The averaged value $S_{\textrm{HST},i}/K$ gives the average similarity between the $i-$th class in the training set and Document HST. The left panel of Figure \ref{fig_similarity} shows the average cosine similarity values normalized by ``sci.spacce" similarity value. We see that the similarity between ``sci.space" and Document HST is the highest, while the similarities between ``comp.windows.x", ``comp.sys.mac.hardware", and Document HST are almost zero. The similarities between seven classes including ``talk.politics.mideast". ``soc.religion.christian" and Document HST are all negative, which indicates that these classes have completely different topics, contexts, or meanings from the Document HST. 

It should be noted that the TF-IDF text embedding technique and PCA relies solely on the documents contained within a dataset, independent of any other models. This approach provides a universal standardized way to measure the similarity between each class in the training dataset and any document, so it can be called as a form of \textbf{"universal text embedding"} methodology. 

\vspace{0.06in}

\subsubsection{\textbf{Model-Based Particular Embedding Method}}

Universal text embedding methods are often computationally expensive. For example, the TF-IDF method  is clearly an expensive method that requires processing the entire training dataset. If the training set is very large, TF-IDF will produce a huge number of dimensions, making dimensionality reduction such as PCA a time-consuming and laborious task (see Section \ref{section_complexity} for more discussion). 

%The above method of vectorizing a specific training dataset using TF-IDF and PCA provides a standardized way to measure the similarity between each class in the training dataset and a particular document. Cosine similarity for each class can be accumulated for any documents. We can introduce a ruleset to map text accumulated cosine similarity values to a scoring system, which can be used to measure the engagement of each class in the sense of text similarity. However, it is clearly an expensive method that requires processing the entire training dataset. If the training set is very large, TF-IDF will produce a huge number of dimensions, making dimensionality reduction a time-consuming and laborious task. 

A more convenient method for calculating text similarity is to use pre-trained models to directly implement text embedding. For example, BERT (\cite{BERT}) provides several methods for vectorizing any text document. For demonstration, we used Sentence-BERT (SBERT, \cite{SBERT}) to embed and vectorize documents in Newsgroup20 training dataset and Document HST. The left panel of Figure \ref{fig_similarity} shows the results provided by SBERT-Basic, which are obviously different from those provided by TF-IDF and PCA. Although ``sci.space" still has the highest similarity with Document HST, other classes also have similar cosine similarities with Document HST. There is no negative similarity between any class and Document HST. This is because SBERT-Basic was pre-trained on the \href{https://nlp.stanford.edu/projects/snli/}{SNLI} dataset, which is a quite different dataset from Newsgroup20.  As a result, the text embeddings generated by SBERT-Basic and those obtained by directly applying TF-IDF and PCA on the Newsgroup20 training dataset are quite distinct from each other.  In addition, we also used text embedding provided by another pre-trained model called \href{https://huggingface.co/sentence-transformers/all-MiniLM-L6-v2}{all-MiniLM-L6-v2}, which was trained on various datasets, to calculate the similarity between each class and Document HST, and although ``sci.space" still shows the highest similarity, it can be seen that it provides different result from TF-IDF and SBERT-Basic. 

The right panel of Figure \ref{fig_similarity}  shows the results of another demo, which is the similarity between the prompt {\ttfamily{\small Why the sky is blue?}} and each class of Newsgroup20. It can be seen that using the TF-IDF and PCA embeddings directly built from the Newsgroup20 training dataset and the embeddings from other two pre-trained models provide significantly different similarity distributions. In Section \ref{section_class} , {\ttfamily{\small Why the sky is blue?}}  has the highest probability for ``sci.space", but in text similarity analysis, the class most similar to this sentence is other class such as ``misc.forsale", which means the results of classification and text similarity may be contrast with each other. 

Clearly, different text embedding methods will provide different results for text similarity. Which method should we use to calculate text similarity and creating a scoring system?

\vspace{0.06in}

\subsubsection{\textbf{Scoring System based on Text Similarity}}

If we aim to set up an alternative prompt-based scoring system using text similarity, there are two feasible approaches. Figure \ref{fig_sim_score} shows the two possible approaches. 

The first approach is to use a universal text embedding technique to vectorize all documents. For example, TF-IDF and dimensionality reduction can be considered as one of such techniques. Moreover, while some text embedding techniques rely on models, we can still use them for text embedding if publicly available pre-trained models gain widespread acceptance among the public. 

The second approach is closely related to the special AI tool we want to build the scoring system for. The AI tool may provide its own text embedding technique, which is created by its own training dataset and model architecture. For example, GPT also provides several pre-trained embedding models, including Ada (1024 dimensions), Babbage (2048 dimensions), Curie (4096 dimensions), Davinci (12288 dimensions), we can use the GPT API to generate text embeddings and convert documents into vectors, calculate the similarity between any documents and GPT, then build a scoring system via text similarity specifically for GPT.  

For any input prompt, the score of text similarity between the prompt and the $i-$th class can be calculated by (similar to equation [\ref{equ_HST}], see also Figure \ref{fig_sim_score}):
\begin{equation}
s_{\textrm{p}i} =   \frac{1}{K}\sum_{j=1}^{K} \frac{\textbf{V}_{\rm p}  \cdot \textbf{V}_{ij}}{|\textbf{V}_{\rm p}||\textbf{V}_{ij}|},\label{equ_sim_prompt}
\end{equation}
where $s_{\textrm{p}i}$ is the cosine similarity score, $K$ is the number of documents in that class, and $\textbf{V}_{\rm p}$ is the prompt embedded as a vector. To sum up $s_{\textrm{p}i}$ for all prompts, we can get the raw score for $i-$th class:
\begin{equation}
s_{i} = \sum_{\rm p} s_{\textrm{p}i}  = \sum_{\rm p} \sum_{j=1}^{K} \frac{\textbf{V}_{\rm p}  \cdot \textbf{V}_{ij}}{K |\textbf{V}_{\rm p}||\textbf{V}_{ij}|},\label{equ_sim_score}
\end{equation}
and similar to equation (\ref{equ_RI}), the revenue shared by the $i-$th class, i.e., the $i-$th data provider can be calculated by
\begin{equation}
\mathscr{R}_{i} = \mathscr{R}_{\rm tot} \times S_i, \label{equ_SI}
\end{equation}
where $\mathscr{R}_{i}$ is the revenue shared by $i-$th provider, and $S_{i}$ is the provider's normalized score calculated by
\begin{equation}
S_i = s_{i}/\sum_{i}s_{i}
\end{equation}

%The first idea is to use a text embedding technique provided by a publicly available pre-trained model that accepted by the public (for example, SBER-Basic or all-MiniLM-L6) to calculate text similarity. The advantage of this approach is that the calculation is clear and does not require knowledge of the special training set from a specific model. So it provide a universal method to calculate text similarity which is independent of various models. However, the disadvantage of this method is also obvious: text embedding heavily depends on the training dataset, and embedding provided by a pre-trained model on n unrelated training set may not reflect the characteristics of the specific training set. 

\begin{figure*}[t] 
 \centering
 \includegraphics[width=0.9\textwidth]{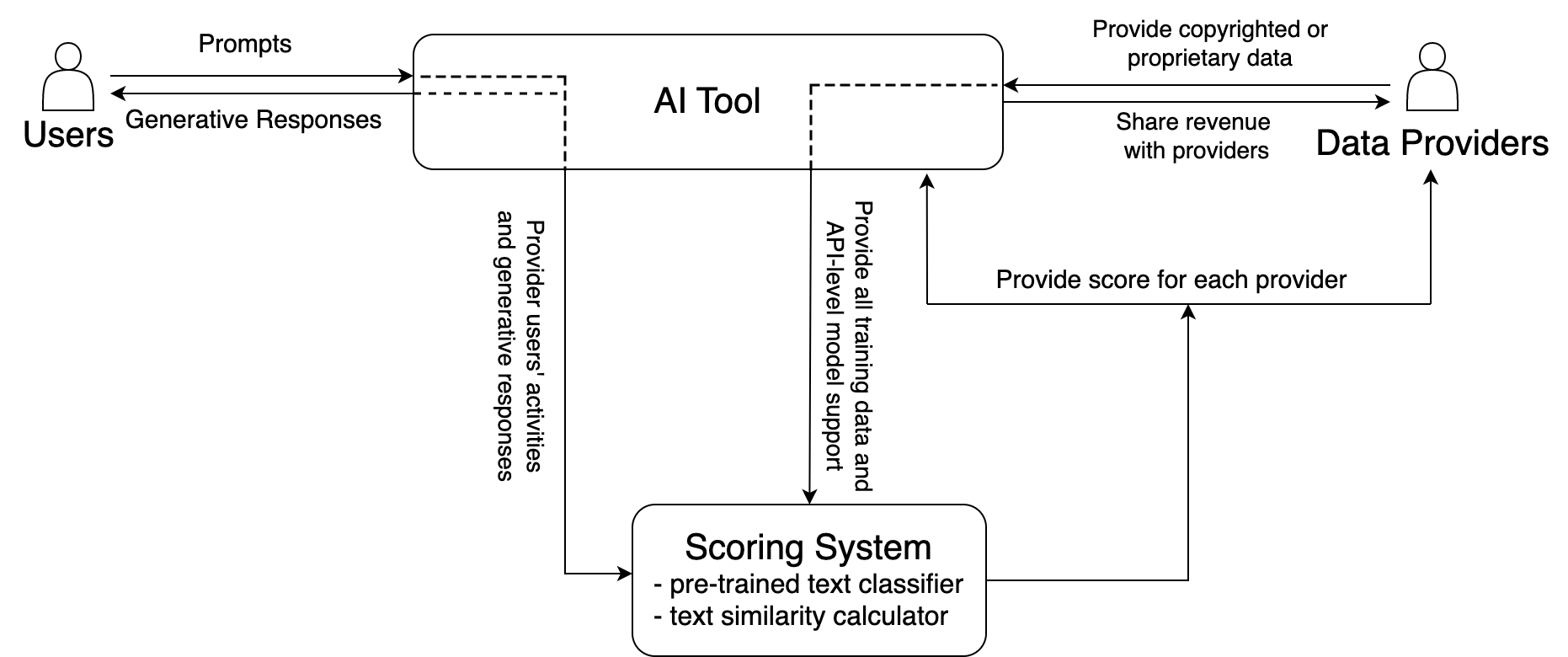}
 \caption{Illustration of the prompt-based scoring system model and the relation between the AI tool, its users, data providers and the scoring system. Users input prompts and receive generative responses from an AI tool. In order to effectively evaluate data providers and allocate revenue accordingly, the AI tool must provide all of its training data with possibly API-level support to the scoring system, so the scoring system can use this information to build a pre-trained text classifier to obtain ``probability score" (Section \ref{section_class}), or text similarity calculator for ``similarity score" (Section \ref{section_text_sim}), while text embeddings are based on model API support. Furthermore, the users' prompts and the AI tool's responses must also be sent to the scoring system, allowing the system to calculate a score for each data provider for each prompt. The sum of these scores across all input prompts will be used to measure the degree of data engagement of each provider. Revenue sharing will then be based on this scoring system, ensuring that data providers are fairly compensated based on their data contributions.  }
  \label{fig_prompt_base_scoring}
\end{figure*}

Keep in mind that, as shown in Figure \ref{fig_similarity}, different text embedding techniques can yield vastly different text similarity computations, and thus result in different scores for each data provider. Currently, there is no general answer as to which text embedding technique is best suited, we can choose the text embedding between universal and model-based methods based on business requirements or other particular purposes. 

Furthermore, when considering not just text embeddings but also more general embeddings such as image embeddings, although the techniques for embeddings and computing similarity can differ, the general approach for establishing a scoring system via document similarity is similar. Section \ref{section_text2image} provides a more in-depth discussion on image embeddings.

Figure \ref{fig_prompt_base_scoring} represents a modified version of Figure \ref{fig_text_classifier}, illustrating the relation between an AI tool, its users, data providers, and a scoring system which is the model for revenue sharing. This figure serves as a summary of Sections \ref{section_scoring_require}, \ref{section_class}, \ref{section_text_sim}. The scoring system is prompt-based, which each data provider's final score being a combination of the pre-trained classifier's probability score and the text similarity calculator's similarity score. It is worth noting that the scoring system is independent of the AI tool/model, although it may utilize the embedding technique provided by the AI tool.

\subsection{Complexity of Scoring Systems}\label{section_complexity}

%In a short summary, for any AI tool, we can rate the data engagement of its various data providers by building multiple scoring systems based on: (1) a pre-trained text classifier, or (2) calculating text similarity with some text embedding techniques.  

Now let us discuss the complexity and cost of the two scoring methods: (1) a pre-trained text classifier, or (2) calculating text similarity with some text embedding techniques. 

The first scoring system requires training a classifier model on the entire training dataset, which can be time-consuming. However, training a tree-based classifier is always less expensive than a semi-supervised deep learning model using the same data size. For example, the preprocessed  training data for GPT-3 is 570 GB, we expect that combination of text embedding and classification training will cost approximately $\sim$\$1000 -\$2000. Once the classifier is completed and becomes a pre-trained model, each prompt input and the response from the AI tool can quickly generate probability scores for all data providers. The complexity of adding up all probability scores given by prompts is linearly related to the number of prompts $\propto O(P)$, where $P$ is the number of prompt inputs. 

The scoring system based on text similarity is more complex. First, if we decide to reduce dimensions after text embedding, the work is more complex than a supervised classification model.\footnote{Consider a tree-based classifier, the time complexity of the training is $\sim O[KFN \log(N)]$, where $K$, $N$ and $F$ are the number of trees, the number of documents, and the number of the features (dimensions) respectively, while the space complexity is $\sim O (KNF)$. On the other hand, the text embedding gives a time complexity of $\sim O(F^3)$, where $F \gg K$ and $F \gg N$ if $F$ is obtained from word count and TF-IDF, which means building text embedding models are much more time consuming  than supervised classification.} For each prompt input and the corresponding generative response, we need to go through all the embedded documents as vectors, to calculate the similarity between each class in the training set and the prompt input with its response. The time complexity of adding up the text similarity provided by a large number of prompts given by equations (\ref{equ_sim_score}) and (\ref{equ_SI}) is $\propto O(NP) \gg O(P)$, where $N$ and $P$ are the number of documents in training set and the number of prompt inputs respectively. For large AI models such as LLMs, the training datasets may have billions or even trillions of documents, and the number of prompts from customers on a daily basis is at least in the billions. Therefore, using text similarity scores to build a scoring system is significantly more complex and expensive than the scoring system based on a classifier.   

However, we can reduce the complexity of text similarity calculation through some methods. For example, we can pre-calculate the characteristic vector for the $i-$th class by
\begin{equation}
\langle \textbf{V}_i \rangle =  \frac{1}{K} \sum_{j=1}^{K} \frac{\textbf{V}_{ij}}{|\textbf{V}_{ij}|}.
\end{equation}
Thus, we have $s_{\textrm{p}i} = \textbf{V}_{\rm p} \cdot \langle \textbf{V}_i \rangle/|\textbf{V}_{\rm p}| $, thereby reducing the computational complexity to $\propto O(C)$ per input, where $C$ is the number of classes. The overall computational complexity would then be $\propto O(CP) \ll O(NP)$.

%\begin{eqnarray}
%\langle \textbf{V}_i \rangle &= &\frac{1}{P} \sum_{j=1}^{K}\sum_{p=1}^{P} \textbf{V}_{ij} \cdot \frac{\textbf{V}_{p}}{|\textbf{V}_{p}|} \nonumber \\
%&= &\frac{1}{P}\sum_{p} \frac{s_{\textrm{p} i}}{|\textbf{V}_{p}|},
%\end{eqnarray} 
%Thus, we have $s_{\textrm{p}i} = \textbf{V}_{\rm p} \cdot \langle \textbf{V}_i \rangle/|\textbf{V}_{\rm p}| $, thereby reducing the computational complexity to $\propto O(C)$ per input, where $C$ is the number of classes. The overall computational complexity would then be $\propto O(CP) \ll O(NP)$.

\begin{figure}[t] 
  \centering
  \includegraphics[width=\linewidth]{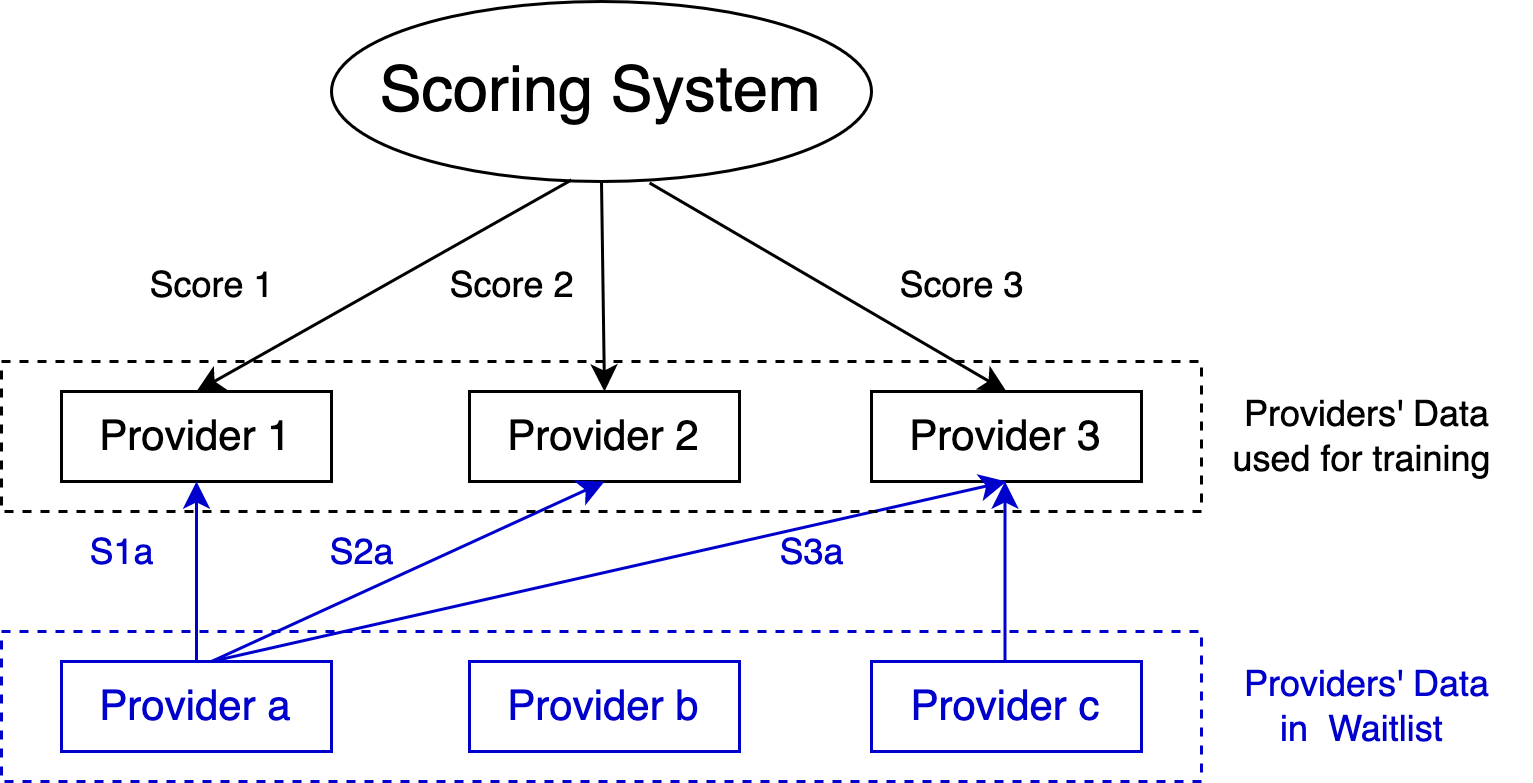}
  \caption{The AI tool's revenue sharing program includes data providers on the waitlist (blue providers) who have joined but whose data has not yet been used for training. These providers can compare their text similarity with providers whose data has been included in the training dataset (black providers), and use equation (\ref{equ_waitlist}) to generate a score.}
  \label{fig_waitlist}
\end{figure}

\subsection{Real-time Scoring System}\label{section_waitlist}

Retrain a large AI model can be challenging and time-consuming. It could take months to collect and preprocess new data, design new model architectures, retrain the model, analyze the model and set up rulesets for the model after retraining. For instance, the first generation pre-trained model of the GPT family, GPT-1, was launched in 2018, while GPT-2 was released in 2019, GPT-3 in 2020, ChatGPT (GPT3.5) in 2022, and GPT4 in 2023. Up until April 2022, the latest data available for ChatGPT is as of 2021. There is no real-time updated large models, at least with current computational capabilities. 

If a data owner is willing to provide data for a large AI tool, it is likely that the provider must wait at least several months even one year until the next training cycle of the AI model, when the provider's data is formally included in the training dataset of the refreshed AI model. Before that, the provider's data is only on a ``waitlist" -- waiting to be used for training. 

In this period of time, can the data providers on the waitlist also share the revenue generated by the AI tool? 

The answer is yes. 

Just like website owners who join Google AdSense can immediately get started to share revenue from Google ads, there is no reason why data providers who have shared data with an AI tool should have to wait until the next training cycle to officially join its revenue sharing program. Here, we provide an idea how to calculate the score of a data provider before their data is officially used. Figure \ref{fig_waitlist} gives the illustration. Assuming the $i-$th class in the formal training dataset has an normalized score of $P_i$, then the score of the $j-$th data provider on the waiting list can be calculated by
\begin{equation}
W_j = \sum_{i}P_{i}S_{ij}, \label{equ_waitlist}
\end{equation}
where $S_{ij}$ is the similarity between the $j-$th provider's data on the waiting list and the $i-$th class in the training dataset, and the scores are normalized by
\begin{equation}
\sum_{i}P_i = 1, \sum_{i} S_{ij} = 1.
\end{equation}
Note that $P_{i}$ is a prompt-based score, so equation (\ref{equ_waitlist}) can also provide a prompt-based score for data providers in the waitlist. If the AI tool is willing to share revenue with the data providers on the waitlist, $W_j$ shows a method for revenue distribution.

\subsection{Third-Party Transfer Learning vs. Centralized Dataset}\label{section_transfer_learning}

Due to data privacy or other reasons, some data owners may want to use a certain AI tool, but are not willing to provide data to the company or organization who owns the AI tool. These data owners may have a small amount of data that is not enough to train a model from scratch, but they can use a technique called transfer learning to leverage the knowledge that a large AI tool has already learned from its massive training data. Transfer learning uses a data owner's special data to fine tune the parameters of the pre-trained AI tool, so that the tool can be adjusted and used for the data owner's specific task \cite{TL_LLM_1, TL_LLM_2}.  

If the performance of transfer learning is good enough, a data owner may be able to leverage the AI tool for a specific task without sharing their own data with the AI tool. In this case, the data owner is obviously not a data provider of the AI tool, and we cannot use the scoring system mentioned above to score their data. So, which is better? Doing transfer learning? Or data owners providing their data to the AI tool and directly obtaining generative responses from the tool? 

There are two scenarios. The first scenario is that data owners remain third-party, and the second scenario is that the centralized AI tool accesses to more and more data and possibly becomes ``big brother"\footnote{Another possibility is to do federated learning for LLMs. We are not going to discuss this scenario in this paper}. From a technical standpoint, the second scenario will achieve at least no worse model performance than the first scenario\footnote{Because after receiving the customer/data owner's data, the AI tool can first do transfer learning for the customer's special task, which ensures that to provide same performance as that done by the customer themselves. Then the AI tool can do something more powerful than the customer alone: it can add the data to its whole training dataset to do various training including semi-supervised and reinforcement learning to further improve the tool's performance on that special task. Meanwhile the general performance of this AI tool can possibly be improved.} So from the perspective of model performance alone, the second scenario is definitely better. Moreover, if a third party data owner does transfer learning alone, they can only benefit themselves,  while sharing data with the AI tool can also benefit everyone who uses the tool. From the perspective of AI utilitarianism, the latter scenario is also better. Therefore, a good plan for a third party data owner is that the owner can submit data to the AI tool to become the tool's data provider, and join the revenue sharing program. While the provider is still on the waitlist (Section \ref{section_waitlist}), it can temporarily use transfer learning technique to obtain generative responses from the AI tool, and when the provider's data is officially used for AI tool's next iteration of training, the provider can obtain better responses from the tool, benefitting from the revenue-sharing program, and also benefitting the public. 

On the other hand, what if the third party has data privacy or other legal considerations preventing them from data sharing, and how to prevent the AI tool which acquires more and more data from becoming a ``big brother", these topics need to be further discussed from a legal perspective, which is beyond the scope of this paper.

%Use your own data, do transfer learning to fine-tune LLM model. And used by you own. 

%On the other hand, share your data with LLM 

\section{Discussion: Revenue-sharing Model for AI Tools other than LLMs}\label{section_discussion}

The above discussion on the revenue-sharing business model is generally applicable to AI tools, but it has been primarily discussed in the context of LLMs, particularly using the GPT family as case studies.  Now let us look into some AI computer vision and multimodal tools. 

\subsection{AI Text-to-Image Generators}\label{section_text2image}

As mentioned in the Introduction (Section \ref{section_intro}), currently there are a couple of popular ``text to image" AI image generators. Dall-E was launched in January 2021, which can be considered as a milestone event \cite{Dall_E}. Although its training dataset was not publicly available, we know that the training data for text-to-image models typically consists of text-image pairs. OpenAI has provided an algorithm called CLIP (Contrastive Language-Image Pre-Training), which can embed both text and images simultaneously \cite{CLIP1}. 

Following Dall-E, the year 2022 could be marked as the rise of AI text-to-image generators. In that year, OpenAI released \href{https://openai.com/product/dall-e-2}{Dall-E 2} in April, \href{https://www.midjourney.com}{Midjourney} launched the first version in July, followed by Stability AI's \href{https://stablediffusionweb.com}{Stable Diffusion} released in August, and Google's \href{https://imagen.research.google}{Imagen} in November. In 2023, new launched AI text-to-image generators so far included Google's \href{https://parti.research.google}{Parti}, \href{https://starryai.com}{Starryai}, \href{https://dream.ai}{Dream by WOMBO}, and so on.  

As of today, many AI text-to-image generators have not publicly opened their training datasets. The training dataset for Dall-E 2 was very likely composed of tens of millions image-text pairs. The more recently text-to-image generators used billions even trillions of image-text pairs. There are various reasons why these companies have not publicly opened the datasets. One reason is that the datasets may contain proprietary data only can be used by particular companies. Another reason may be that those companies are concerned about the potential for misuse of the data. There is also a reason that companies may believe that keeping the training dataset private gives them a competitive advantage over others.

Stable Diffusion is one of the few text-to-image generators that has transparent model with opened training dataset, which was taken from LAION-5B, a publicly available dataset derived from Common Crawl consisting of 5.85 billions CLIP-filtered pairs of images and caption \cite{LAION-5B}. The dataset was created by \href{https://laion.ai/projects}{LAION projects}, and Stable Diffusion was trained on at lease three subsets of LAION-5B including laion-high-resolution and laion-aesthetics v2 5+. An analysis conducted by a third party, which indexed a sample of 12 million images, revealed that almost half of the images ($\sim47\%$) were obtained from around 100 domains. The largest number of images came from Pinterest, followed by user-generated content platforms such as WordPress, Smugmug, Blogspot, Flickr, DeviantArt, Wikimedia Commons and Tumblr. Of the top 25 artists in the dataset, only three are still living\footnote{The third party analysis is from \href{https://waxy.org/2022/08/exploring-12-million-of-the-images-used-to-train-stable-diffusions-image-generator}{Exploring 12 Million of the 2.3 Billion Images Used to Train Stable Diffusion’s Image Generator}, and the 1.2 million images with domains can be seen here \href{https://laion-aesthetic.datasette.io/laion-aesthetic-6pls/images}{laion-aesthetic-6pls}.}. 

Here is an important issue: whether AI image generators use copyrighted images in their training datasets without the owners' permission? Although many image generators' training datasets are still a black box, people can still use some methods to check the similarity of images and detect if their copyrighted work has been used for training. An interesting website called ``Have I been trained"\footnote{This is the link of this tool: https://haveibeentrained.com} is designed to check whether your artwork has been included in the LAION-5B dataset. Currently, AI image generators, as a new technology, have already faced some legal disputes. As mentioned in Section \ref{section_intro}, a group of artists have sued Stability AI, Midjourney and DeviantArt for copyright infringement. In addition, Getty images has filed a lawsuit against Stability AI, accusing Stable Diffusion of ``brazen infringement of Getty Images' intellectual property on a staggering scale" and misusing more than 12 million Getty photos to train its Stable Diffusion AI image-generation system\footnote{See the report \href{https://www.theverge.com/2023/2/6/23587393/ai-art-copyright-lawsuit-getty-images-stable-diffusion}{Getty Images sues AI art generator Stable Diffusion in the US for copyright infringement}.}. Faced with the issue of copyright infringement, AI image generator companies have emphasized that they will comply with the Digital Millennium Copyright Act (DMCA) and protect the copyright of image owners.

Meanwhile, the U.S. Copyright Office has taken the position that AI-generated images do not qualify for copyright protection, as they are not the result of human authorship and therefore do not meet the definition of originality\footnote{US Copyright Office decided that AI generated works are not eligible for copyright, see the \href{https://www.federalregister.gov/documents/2023/03/16/2023-05321/copyright-registration-guidance-works-containing-material-generated-by-artificial-intelligence}{statement of policy}, but AI-assisted work may still be copyrighted.}. We hope that AI image generators will be better regulated in the near future. To ensure that every artist and copyrighted work to be respected, AI image generators must to be more transparent to disclose their training datasets. However, at the same time, we also expect the revenue-sharing business model discussed earlier based on LLMs could to be applied to AI image generators. As discussed in Section \ref{section_utili}, the connection between human artists, copyright owners, and AI image tools should not be viewed as a hostile, zero-sum game but rather as a collaborative and mutually beneficial relationship.

Similar to the revenue-sharing business model of LLMs, we can also establish a scoring system for image providers of AI image generators, which would be based on the combination of some image classification models and image similarity calculators. In principle, as long as we replace ``text" with ``image" in Figures \ref{fig_text_classifier}, \ref{fig_sim_score}, \ref{fig_prompt_base_scoring} from Section \ref{revenue_share_model_AI}, we can use some image embedding techniques to build an image classifier, and image similarity calculator on the image training dataset for a certain AI image generator\footnote{There are many techniques to embed images and check image similarities. For simplicity I will not go into detail about it in the main text. For example, image embedding techniques include CNN (Convolutional Neural Networks) such as \href{https://pytorch.org/vision/stable/models/resnet.html}{ResNet50}, \href{https://pytorch.org/vision/stable/models/inception.html}{Inception V3}, \href{https://pytorch.org/vision/stable/models/efficientnet.html}{EfficientNet}, etc, Autoencoders Embeddings, ORB (Oriented FAST and Rotated BRIEF), CLIP, and so on. There are also various ways to calculate image similarities. For example, FID (Frechet Inception Distance) is a widely used method, or we can just simply use cosine similarity after image embeddings. Use which embedding technique and what similarity calculator can be based on the AI tool which scoring system is established for, or a universal/standard way for multiple AI tools (see Figure \ref{fig_sim_score}). We will do a series of experiments in the future based on various embedding techniques and similarity calculators, but the main goal of this paper is to discuss the feasibility of the revenue-sharing model for AI tools.}, and establish a scoring system to measure the degree of engagement for each image provider, or even individual images. Different from LLMs, image owners/providers may be more concerned about the engagement of their individual artworks in the image generator, rather than just the overall engagement of their all works. Additionally, rather than turning each image provider into a class to build an image classifier, it may be more reasonable to classify images by art style, genre, topic rather than by image providers. Therefore, the image classifier may be multi-tasking\footnote{For MVP (minimum viable product) we still want to set each image provider as one class to build the classifier, which is consistent with text classifier in Section \ref{section_class}. The MVP will be provided in the near future. }.

\begin{figure*}[t] 
  \centering
  \includegraphics[width=0.8\textwidth]{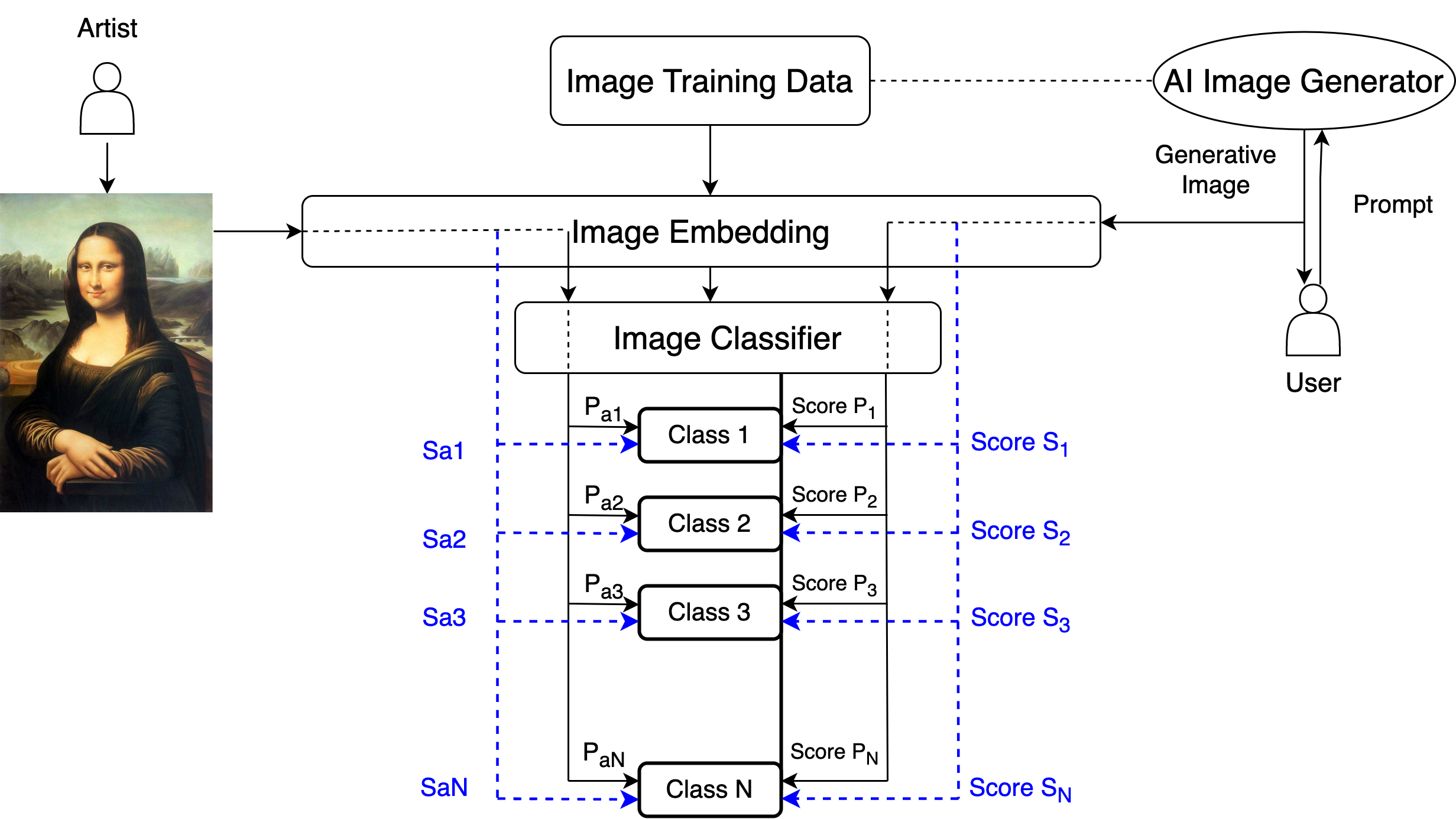}
  \caption{The prompt-based score system is used by image providers for an AI image generator, including two sub-systems: the system based on image classification and probability scores (black lines under ``Image Classifier"), and another system based on image similarities (blue lines under ``Image Embedding"). Each artwork/image is given a score to highlight its contribution. To train a particular AI image generator, we can utilize an image embedding technique (e.g., CLIP) to convert images into vectors, and then build an image classifier based on all images in the training dataset, where the classes can be based on image providers, artistic styles, themes, genres, and so on. Suppose an artist provides an artwork (Mona Lisa in this Figure) to the training dataset, the image classifier gives a series of probability scores for this artwork ($P_{a1}, P_{a2}, P_{a3}$...), and the averaged similarity between each class and the artwork ($S_{a1}, S_{a2}, S_{a3}$...). On the other hand, for a user input prompt, an image is generated based on this prompt. We can calculate the probability score set contributed by this generative image ($P_{1}, P_{2}, P_{3}$...), and a similarity scores also ($S_{1}, S_{2}, S_{3}$...). The final score of the artwork can be calculated based on these scores (see Equation [\ref{equ_image_score}] ), and is still prompt-based. }
  \label{fig_image_rev}
\end{figure*}

Figure \ref{fig_image_rev} shows the scoring system that AI image generators can reply on for their revenue-sharing business model. In general, image scoring system is very similar to the text scoring system discussed in Section \ref{revenue_share_model_AI}, but we highlight individual artwork by an artist in Figure \ref{fig_image_rev}, and provide a method for measuring the importance of that particular artwork. Assuming the images used for a certain AI image generator can be divided into $N$ categories to build an image classifier, the probability scores of an artist's artwork (e.g., Mona Lisa in Figure \ref{fig_image_rev}, hereafter ML) for each class of the image classifier are $\{P_{ai}\}$ $(i =1,2,3 ... N)$, the averaged image similarity score between each class and the artwork (ML) is $\{S_{ai}\}$. For one generative image created by the AI image generator for a user's text prompt, the probability scores of all classes for this generative image are $\{P_{ij}\}$, and the averaged similarity with each class is $\{S_{ij}\}$, where j means the score sets are prompt-based and for the p$-$th prompt. Here we assume all score sets are normalized. Then, based on the image classifier, the probability score of this artwork (ML) is calculated by:
\begin{equation}
P_{\rm ML, p} = \sum_{i=1}^{N}P_{ai} P_{i \rm p},
\end{equation}
and the similarity score of ML is 
\begin{equation}
S_{\rm ML,p} = \sum_{i=1}^{N}S_{ai} S_{i \rm p},
\end{equation}
And the final score of this artwork ML is calculated by
\begin{equation}
P_{\rm ML} = \sum_{\rm p}\sum_{i=1}^{N}P_{ai} P_{i \rm p}, \quad S_{\rm ML} = \sum_{\rm p}\sum_{i=1}^{N}S_{ai} S_{i \rm p} \label{equ_image_score}.
\end{equation}

Note that image similarity scores are more straightforward if the image embedding method has been selected. For image classifier, if the classifier is multi-task (for example, Mona Lisa can be multi-labeled by Renaissance, oil painting style, Da Vinci, female artwork, etc.), we can change probability scores for each class $\{P_{ai}\}$ and $\{P_{i}\}$ mentioned above be a set of two-dimension scores $\{\textbf{P}_{ai}\}$ and $\{\textbf{P}_{i}\}$\footnote{i.e, each $\textbf{P}_{ai}$ and $\textbf{P}_{i}$ are in the form of [probability\_score\_belong\_to\_one\_class, probability\_score\_not\_belong\_to\_one\_class], where the sum of the pair scores is one.}. and calculate the scores as 
\begin{equation}
P_{\rm ML} = \sum_{\rm p}\sum_{i=1}^{N}\textbf{P}_{ai} \cdot \textbf{P}_{i \rm p} \label{equ_image_score_2}.
\end{equation}

Although the U.S. Copyright Office has taken the position that AI-generated images do not qualify for copyright protection, we can provide provide a method to quantify copyright. Actually, we can use the revenue-sharing scoring system to allocate the contributions of human artists, the AI tool, and user who generates an AI-generated image, and allocate the proportion of copyright accordingly, so copyright can be quantified.

\subsection{AI in Healthcare}\label{section_healthcare}

AI may potentially have a significant impact in the healthcare field as well. The most promising applications of AI in healthcare include medical imaging, predictive analytics, drug development, personalized patient care, and so on.

However, the development of AI in healthcare has not been as fast as in other fields. For example, the well-known IBM \href{https://www.ibm.com/watson-health}{Watson Health} and \href{https://health.google}{Google Health} both encountered challenges. IBM Watson Health was founded in 2015 with the goal of using NLP and machine learning to ``transform healthcare with cognitive capabilities". However, IBM Watson Health faced significant challenges soon after. In 2018, IBM Watson Health underwent a massive layoff, and eventually IBM sold Watson Health in 2022. Google launched Google Health in 2008, and a new division called Google Health AI in 2018. Similar to IBM Watson Health, Google Health also encountered a series of problems and setbacks, leading to the dissolving of Google Health as single unified division and downgrading of it to Google Research\footnote{The 2018 Watson Health layoff news can be seen \href{https://www.axios.com/2018/05/31/ibm-watson-health-confirms-layoffs-1527800584}{here}, and IBM sold Watson Health news \href{https://www.hpcwire.com/2022/01/21/ibm-watson-health-finally-sold-by-ibm-after-11-months-of-rumors}{here}. The Google shut down Google Health division news in 2021 \href{https://www.fiercehealthcare.com/tech/google-dissolved-its-unified-health-division-what-s-next-for-its-health-tech-strategy}{here}.}. 

Overall, healthcare AI faces several big challenges \cite{healthAI_1}. Firstly, how to build trust between doctors/patients and AI tools? Explainable AI may help to make AI tools less of a black box, but there are still debates and criticisms about the role of Explainable AI in the medical realm \cite{XAI_health}. Secondly, the lack of sufficient data for AI tools in healthcare has limited their performance. In healthcare, collecting, managing and sharing data is complicated and difficulty due to privacy concerns, regulations, technical constraints, cost and disincentives \cite{healthAI_2}. Obtaining high-quality data user regulatory constraints is still a significant challenge. 

The revenue-sharing business model for AI tools may have significant potential in other fields, but could we also introduce a revenue-sharing model to encourage patients to share their records and gather more data for model training? This is a question worth exploring. 

%One example of synthetic data improving model performance can be found in the domain of medical imaging. In medical imaging, it can be challenging to obtain a large enough dataset due to privacy concerns and other limitations. This can lead to overfitting and poor generalization of machine learning models trained on these small datasets. To overcome this issue, researchers have used synthetic data to augment their training datasets.

\subsection{Multimodal AI Tools}\label{section_multimodal}

OpenAI's GPT-4 is a large multimodal model. We can foresee that there will be more and more multimodal AI tools in the future. We will not only have tools for text-to-text, text-to-image, image-to-text, text-to-music, text-to-video tools, but also hybrid tools, for example, a tool to input both text and image, and output text, music, and video simultaneously. For multimodal tools, how can we establish a revenue-sharing scoring system?

Following the discussion in Section \ref{section_metrics}, we want to ensure that the revenue sharing for multimodal tools is still based on prompts. One approach is to examine the generative content of the AI tool, classify it into different modalities, build a scoring system for each modality, and combine scores together so every text/image/audio/video provider will be assigned a prompt-based score. For example, if the generated content contains both text and images, we can use the methods discussed in Sections \ref{section_class} and \ref{section_text_sim} to score each provider in the text training dataset and the method in Section \ref{section_text2image} to score each provider in the image training dataset. This approach to measure the engagement of data providers is called ``generative content-based approach".

%Finally, we can sum up the scores generated by all generative content to obtain scores for all data providers of the multimodal tool.

%\section{Requirements for AI Companies}

\vspace{0.1in}

\section{Conclusion and Future Work}\label{section_conclusion}

ChatGPT has created a significant impact in the field of artificial intelligence (AI), heralding a new era in the development and application of AI. We can foresee that exceptional AI tools will soon reap considerable profits. However, we must ask a crucial question: should AI tools share revenue with their data providers? The short answer is: Yes.

For deep learning models, especially large models such as Large Language Models (LLMs), we need to acquire more and better quality data to achieve better model performance. Here "more and better data" not only means collecting a large training dataset, but also has at lease four meanings: (1) Enhancing the diversity and amount of training data and reduce the data noise can improve models’ learning ability, and decrease the chances of errors. (2) Using data specific to a certain task of domain knowledge can fine tune the model for that task or domain. (3) Keep training data up-to-date to produce more precise and recent results. (4) Incorporating more human feedback to establish better ground truth. 

However, due to data privacy and copyright laws, most AI tools can only collect training datasets in the public domain. Some AI tools may have access to certain copyrighted data, but their access could be limited. On the other hand, some other AI tools may have already breached copyright laws, which can result in legal issues. Even if the data of AI tools is currently in the public domain, it is highly likely that the some original data owner will lose more and more visitors and customers, and possibly refuse AI tools to use their data. The game between data owners and AI tools look like a zero-sum game. In this paper, we discuss a better way for AI tools to obtain data and benefit all parties, and to establish a new revenue-sharing business model for building a good AI ecosystem in the upcoming AI era.

Taking LLMs as an example, in the fierce competition among various models and tools, a successful LLM must have: (1) more and better quality data than other tools, (2) more powerful computing resources, and (3) faster iterations to update the model and to meet customer needs. Just as today's search engine market is dominated by Google search with over 90\% market share, one or several best LLMs with their chatbots could potentially dominate the language tool market or even multimodal market in the future. This is called "LLM monopoly," which could pose a great threat to various text data owners. To establish a utilitarian AI era where AI tools can keep improving performance and a wide range of data owners can benefit from it, a revenue-sharing model established by AI tools is necessary.

Of course, establishing a revenue sharing model for AI tools is very challenging. Today's revenue sharing models, such as Getty Images'  royalty-based model and Google AdSense's cost-per-action model in the Web 2.0 era, will not work in the AI era. The existing revenue sharing models based on old metrics such as clicks, page views, and conversion rates (Figure \ref{fig_google_model}) will be replaced by new metrics-based business models such as prompts, cost per prompt and data engagement degree (Figure \ref{fig_share_rev_model}). In order to fairly allocate revenue with data providers, AI tools must establish a scoring system to evaluate each data provider. There are several crucial prerequisites for this scoring system: (a) The scoring system must be based on prompts. (b) The scoring system should be simple and easily explainable to the general public. Therefore, it is not advisable to use deep neural networks to build such a system. (c) The AI tool's entire training dataset should be open and transparent, allowing third-party verification. (d) The scoring system must be as independent from the AI tool as possible. The AI tool itself can provide API-level information to the scoring system, but we need to ensure that if we treat the AI model as a black box, the scoring system can still be established and explainable.

Specifically, for an AI tool, we can establish a supervised classification model for data providers in its training dataset (Section \ref{section_class}). If we consider each data provider as an independent class, or a combination of data providers as one class, we can build a pre-trained classification model on this training dataset based on the class (provider) to which each document belongs. Once a user inputs a prompt and receives a generative response from the AI tool, we can use the pre-trained classifier to assign each class a probability score based on the prompt or the response, so the probability scores reflect the degree of engagement of each class for this prompt. The summation of probability scores for all prompts can be used to establish a scoring system. Figure \ref{fig_text_classifier} shows the basic pipeline for establishing a scoring system based on a pre-trained text classifier. In Section \ref{section_class}, we also provide demonstrations on how to establish a text classifier and its corresponding scoring system using benchmark text datasets.

Building a scoring system based on probability scores of a text classifier on the training dataset of an AI tool can be comprehensible, but its interpretation could be challenging if the training data is incomplete or poorly labeled. On the other hand, for language tools we can score data providers by calculating text similarity. Specifically, we can use a universal or tool/model-based particular text embedding technique to convert documents in the training dataset into vectors, and calculate the averaged text similarity between each prompt or the generative response and each data provider (Section \ref{section_text_sim}). Each data provider can obtain a prompt-based text similarity score, which can be summed up prompt by prompt, so we can obtain a final score for each data provider, thus establishing a scoring system based on the the text similarity calculator (see Figure \ref{fig_sim_score}). 

Note that different text embedding methods will calculate different text similarity scores. Currently, there is no general answer as to which text embedding technique is best suited, so we can choose the text embedding method between universal and model-based methods based on business requirements or other particular purposes. Additionally, we discussed the complexity of different scoring systems, and the feasibility of establishing a scoring system in terms of computational complexity and cost (Section \ref{section_complexity}). Moreover, for data providers that have not yet been included in the training dataset but have already joined the revenue-sharing program, we can also establish a temporary rating mechanism for them (Section \ref{section_waitlist}). 

In Section \ref{section_text2image}, we discussed how to establish a scoring system for AI image generators and calculate engagement rate for individual artworks (see also Figure \ref{fig_image_rev}). Although the U.S. Copyright Office has taken the position that AI-generated images do not qualify for copyright protection, we can provide provide a method to quantify copyright. We can calculate the contributions of human artists, the AI tool, and user who generates an AI-generated image, and allocate the proportion of copyright accordingly, so copyright can be quantified.

The revenue-sharing model used for LLMs can be applied to other AI tools as well. For instance, AI image generators can use a scoring system based on image classification models and image similarity calculators to establish revenue sharing with their image providers. However, different from LLMs, image providers may be more interested in the engagement of their individual artworks rather than their overall portfolio. Furthermore, it may be more appropriate to classify the image database by art style, genre, and topic rather than by image providers. 

Finally, we discussed AI in healthcare (Section \ref{section_healthcare}) and multimodal models (Section \ref{section_multimodal}). Perhaps we could also establish a revenue-sharing model for AI in healthcare, but this issue requires careful consideration. As for multimodal models, which are expected to become more prevalent in the future, we can establish a multimodal scoring system to score data providers for each modality.

Regarding making AI tools share revenue with their providers, building scoring system models for revenue-sharing is just a technical issue, but not the major blocker. The major blocker is how to persuade people to share data, and how to persuade AI tools to share revenue. I have repeatedly argued from the perspective of utilitarianism that establishing a revenue-sharing mechanism would be a mutually beneficial solution for all parties involved. It would facilitate the development of a virtuous cycle that drives forward AI technology and builds an ideal AI ecosystem that benefits everyone. This paper mainly discussed the necessity and feasibility of revenue sharing between AI tools and their data providers,  and provided high-level approaches for building a scoring system for revenue sharing. For practical demonstrations, I used the benchmark text datasets Newsgroup20 and Reuters-21578. To further demonstrate the feasibility of building a scoring system for real data providers, future work should involve using larger datasets that are comparable in size to the training datasets of LLMs to enable more realistic modeling. In particular, for text data, the Pile dataset, which comprises 800GB of diverse text \cite{Pile}, can be employed to train a text classifier and calculate text similarity using various text embedding techniques. For image data, we can use LIOAN-5B dataset to build an image classifier and calculate image similarity using different image embeddings. In order to explain the classifier built by supervised classification, we need to use some explainable AI techniques to better explain the classifiers to laymen. Similar efforts have been made to compute image similarity for LIOAN-5B (as discussed in Section \ref{section_text2image}), but we can enhance our approach by creating a scoring system and assigning scores to each data provider through simulated prompts. In this way, the work of building a scoring system on large datasets can be directly applied to various current and future AI tools.

% \section*{ACKNOWLEDGMENT}
% Omitted for blind reviewing
% \noindent This work is conducted at the center of Data Science and Systems Complexity (DSSC) and sponsored by a  Marie Sk{\l}odowska-Curie COFUND grant, agreement no. 754315.

\bibliography{references}

\bibliographystyle{IEEEtran}

\end{document}